\documentclass{article}
\usepackage[final,nonatbib]{neurips_2024}
\usepackage[utf8]{inputenc} 
\usepackage[T1]{fontenc}    
\usepackage{hyperref}       
\usepackage{url}            
\usepackage{booktabs}       
\usepackage{nicefrac}       
\usepackage{microtype}      
\usepackage{xcolor}         
\usepackage{listings}%
\usepackage{colortbl}
\usepackage{makecell}
\usepackage{subfigure}
\usepackage{graphicx}%
\usepackage{multirow}%
\usepackage{amsmath,amssymb,amsfonts}%
\usepackage{amsthm}%
\usepackage{mathrsfs}%
\usepackage{textcomp}%

\newcommand{\aka}{\emph{a.k.a.},\ }
\newcommand{\et}{\emph{et al.}\ }

\newcommand{\qfnu}{{\includegraphics[scale=0.035]{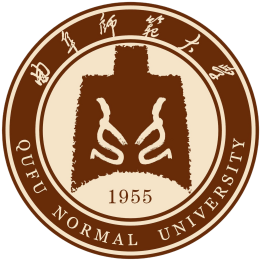}}}
\newcommand{\hit}{{\includegraphics[scale=0.035]{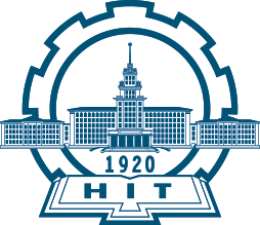}}}

\title{Searching a Lightweight Network Architecture for Thermal Infrared Pedestrian Tracking}

\author{
Wen-Jia Tang$^{\qfnu}$, Xiao Liu$^{\qfnu}$, Peng Gao$^{\qfnu}$, Fei Wang$^{\hit}$, Ru-Yue Yuan\\
$^{\qfnu}$Qufu Normal University \quad
$^{\hit}$Harbin Institute of Technology Shenzhen \\
}

\begin{document}

\maketitle

\begin{abstract}
Manually-designed network architectures for thermal infrared pedestrian tracking (TIR-PT) require substantial effort from human experts. AlexNet and ResNet are widely used as backbone networks in TIR-PT applications. However, these architectures were originally designed for image classification and object detection tasks, which are less complex than the challenges presented by TIR-PT. This paper makes an early attempt to search an optimal network architecture for TIR-PT automatically, employing single-bottom and dual-bottom cells as basic search units and incorporating eight operation candidates within the search space. To expedite the search process, a random channel selection strategy is employed prior to assessing operation candidates. Classification, batch hard triplet, and center loss are jointly used to retrain the searched architecture. The outcome is a high-performance network architecture that is both parameter- and computation-efficient. Extensive experiments proved the effectiveness of the automated method.\\
\textbf{Keywords:} Thermal infrared, pedestrian tracking, machine learning, neural network
\end{abstract}

\section{Introduction}\label{sec:1}

Thermal infrared pedestrian tracking (TIR-PT) is an active research field in computer vision and draws increasing interest in autonomous driving and underwater vehicles. It is generally treated as an object detection problem that intends to follow a specific pedestrian in TIR video sequences~\cite{yuan2022recent,yuan2023thermal}.

Convolutional neural networks (CNNs) have been widely used in TIR-PT to extract features. Most of the CNNs of TIR-PT adopt typical networks such as AlexNet~\cite{alexnet} and ResNet~\cite{resnet} as the backbone architecture, then integrate more layers to use deep features.For example, MMNet~\cite{mmnet} implemented a multi-task matching network based on AlexNet to learn object-specific discriminative and fine-grained correlation feature maps of TIR pedestrians. LMSCO~\cite{lmsco} integrated the motion information of the pedestrian into VGGNet to overcome the background clutter and motion blur of the TIR image. ASTMT~\cite{astmt} designed an aligned spatial-temporal memory network based on ResNet to take advantage of learning scene information for pedestrian localization. The well-performance handcrafted neural networks require substantial effort from human experts. However, these backbones are initially designed for image classification and object detection and thus have many redundant parts when transferred to different tasks. Since TIR-PT is an instance-level detection task, we hope to find a specific architecture suitable for it.

Neural architecture search (NAS) has recently attracted wide attention in classification~\cite{wang2023fp}, detection~\cite{liang2022cbnet}, tracking~\cite{gao2023efficient}, and segmentation~\cite{wang2023automatic}. It is an important advance that automates neural network designing. In this paper, we attempt to search for an efficient network architecture for the TIR-PT task automatically. Given a search space, NAS methods aim to search a high-performance architecture with specific search strategies. Conventional NAS methods cost intensive computation and memory. For instance, the reinforcement learning (RL)-based NASNet proposed by Zoph \et~\cite{zoph2018learning} takes 2000 GPU days to search and evaluate neural networks. AmoebaNet~\cite{real2019regularized} is a well-known regularized evolution method. It costs 3150 GPU days to find a suitable architecture.

\begin{figure}[t!]
    \begin{center}
        \includegraphics[width=0.9\linewidth]{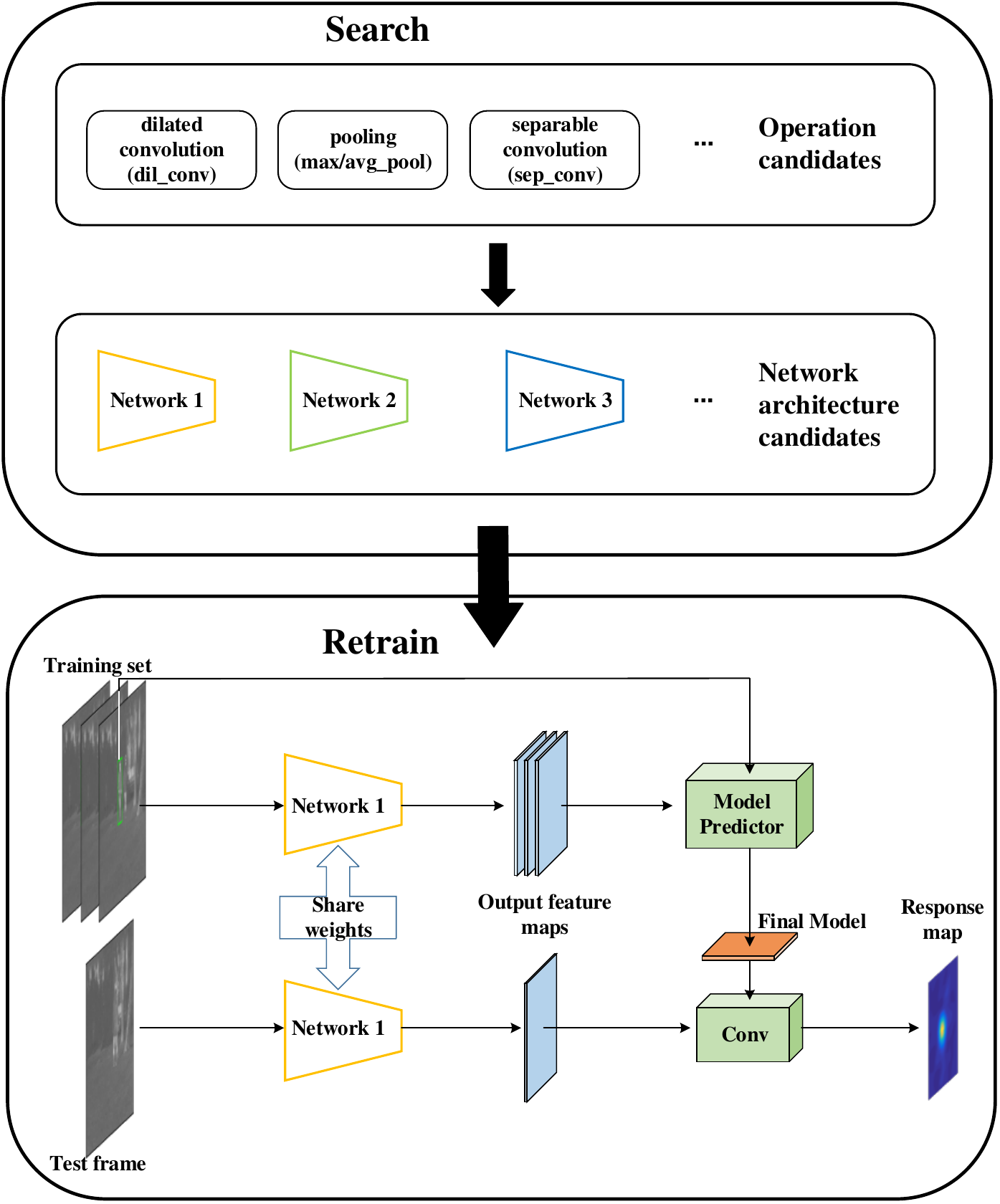}
    \end{center}
    \caption{Overview of the proposed TIR-PT method.}
    \label{fig:arch}
\end{figure}

Inspired by the aforementioned works, in this paper, we employ the discriminative tracking model DiMP~\cite{dimp} as our baseline and use a differentiable searching manner inspired by \cite{gao2023efficient} to learn a preferable network architecture for TIR-PT tasks. We relax the discrete search space to be continuous to optimize the architecture by gradient descent, and design a two-stage (search and retrain) method for TIR-PT, as shown in Fig.\ref{fig:arch}. In the first stage, operation candidates in the search space are used to discover the basic architecture cells (or blocks). Then, the overall network architecture is constructed by stacking cells. Similar to the attention mechanism, apart from updating normal operation (such as convolutions) parameters $\omega$, we also need to update architecture parameters $\alpha$, an attention matrix representing the importance of each operation. Since it is a bi-level optimization problem, $\alpha$ and $\omega$ can be thus updated alternately. Though the differentiable method is relatively faster than other NAS, it still costs a lot of storage. Therefore, we randomly select channels and feed them into operation candidates to reduce computation, as suggested by \cite{xu2019pc}. The efficiency of the search process can be improved in this way. As for the second stage, we chose the most critical operation to form the final network architecture, similar to network pruning. We use joint supervision of classification, batch hard triplet, and center loss to learn more discriminative features to retrain the searched network architecture. Our contributions can be summarized as follows.

\begin{itemize}
  \item We made an early attempt to search automatically for an effective network architecture for the TIR-PT task. The architecture learned on the LSOTB-TIR dataset can be directly transferred to the PTB-TIR dataset with high performance.
  \item We construct single-bottom and dual-bottom cells as basic units, and stack them together to search a network architecture.
  \item Instead of calculating all channels, we randomly select a part of the channels and feed them into operation candidates to accelerate the search process.
  \item Considering the challenging scenarios of TIR-PT, we explore joint supervision of classification, batch hard triplet, and center loss to retrain the searched network architecture.
  \item Extensive experimental results demonstrated the effectiveness and efficiency of our proposed method.
\end{itemize}

The remainder of this paper is organized as follows. Relevant TIR-PT and NAS methods are introduced in Section~\ref{sec:2}. Section~\ref{sec:3} details our methodology for searching for an optimal neural network architecture for TIR-PT. In the following section, Section~\ref{sec:4}, experimental results and discussions demonstrate that the proposed method is efficient for TIR-PT. Finally, a brief conclusion is provided in Section~\ref{sec:5}.

\section{Related works}\label{sec:2}

\subsection{Thermal infrared pedestrian tracking}

TIR-PT represents a significant advancement in surveillance, security, and urban management technologies. Its ability to operate in low-light conditions, independence from weather constraints, and privacy-sensitive imaging make it a versatile tool in various applications. Early TIR trackers used handcrafted features \cite{dense}, but the DSST-tir tracker showed that deep features are more effective \cite{dssttir}. Consequently, more TIR trackers started incorporating deep features. MCFTS \cite{mcfts}, for instance, uses VGGNet layers for an ensemble tracker, while HSSNet \cite{hssnet} transfers a deep Siamese framework from a large video dataset to the TIR domain. Recent advancements include MLSSNet \cite{mlssnet}, which excels in distinguishing distractors, and MMNet \cite{mmnet}, which innovates dual-level deep representation learning. Prevailing backbone networks utilized by existing TIR-PT methods have been designed manually by human experts, which is time-consuming and requires a lot of prior knowledge from experts. Our goal is to automatically search an architecture for the TIR-PT task.

\subsection{Network architecture search}

In recent years, NAS has attracted growing attention and plays a vital role in automatic machine learning (AutoML). The goal is to search for architectures automatically to replace conventional manual ones. Early approaches take a large amount of computation. Zoph \et \cite{zoph2016neural} use a recurrent neural network (RNN) to encode the network architecture, which takes 800 GPUs for 28 days, resulting in 22,400 GPU hours to search for a suitable architecture on the CIFAR-10 dataset. Further, NASNet, proposed by Zoph \et \cite{zoph2018learning}, uses 500 GPUs across four days, resulting in 2,000 GPU hours. Metacontroller is implemented as an RL agent in \cite{cai2018efficient}. It learns to take action for network transformation (widen layer, insert layer, add skip-connections, etc.) to explore the architecture space. ENAS, proposed by Pham \et \cite{pham2018efficient}, speeds up NAS by sharing parameters among child models during the search procedure. Some methods try to reduce computation costs. In particular, Liu \et \cite{liu2018darts} use a differentiable method to relax the discrete operations into continuous variables. The essential edges with high attention-weighted scores will be left during inference. Xu \et \cite{xu2019pc} explore larger batch sizes for either speedup or higher stability with partial-connected mechanism.

Our method is closest to DARTS \cite{liu2018darts}. For RL-based and evolutionary-based methods, feedback (\aka reward) is obtained after a long training. In contrast, feedback (\aka loss) in the gradient-based method is given in every iteration, which is more instant. Therefore, the optimization of the gradient-based method is potentially more efficient. The discrete search space can be transferred to a continuous matrix with softmax, enabling the search to progress end-to-end.

\section{Methodology}\label{sec:3}

\begin{figure}[t!]
    \begin{center}
        \includegraphics[width=\linewidth]{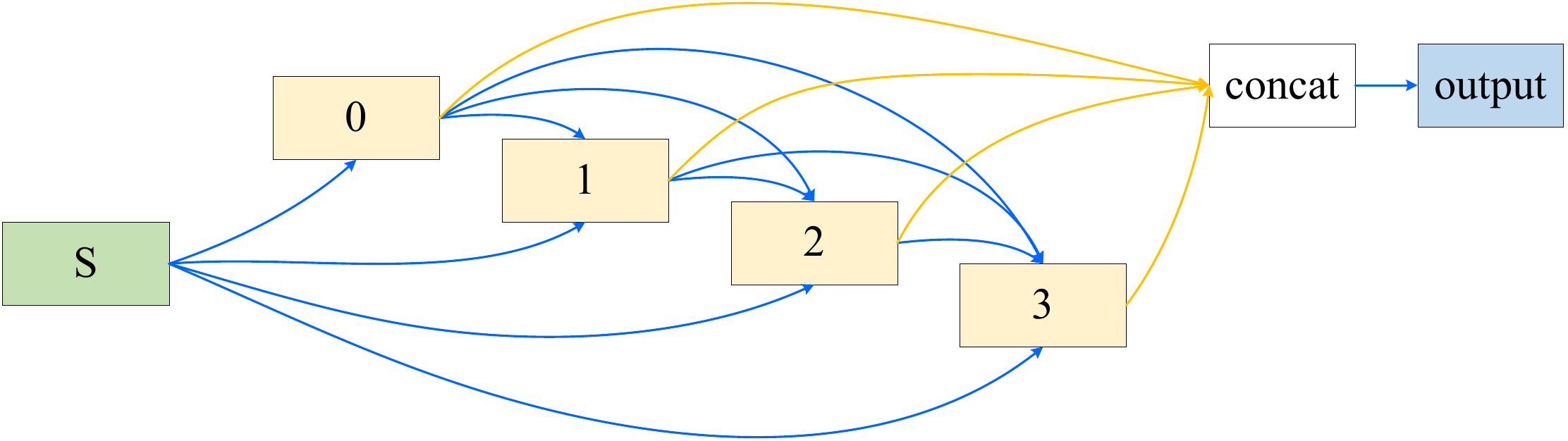}
    \end{center}
    \caption{A single-bottom cell contains six nodes. Node S is the input feature map. Nodes 0, 1, 2, 3 are intermediate feature maps. The last output node is depth-wise concatenation of four intermediate nodes. Each blue edge denotes eight operation candidates.}
    \label{fig:single}
\end{figure}

\begin{figure}[t!]
    \begin{center}
        \includegraphics[width=\linewidth]{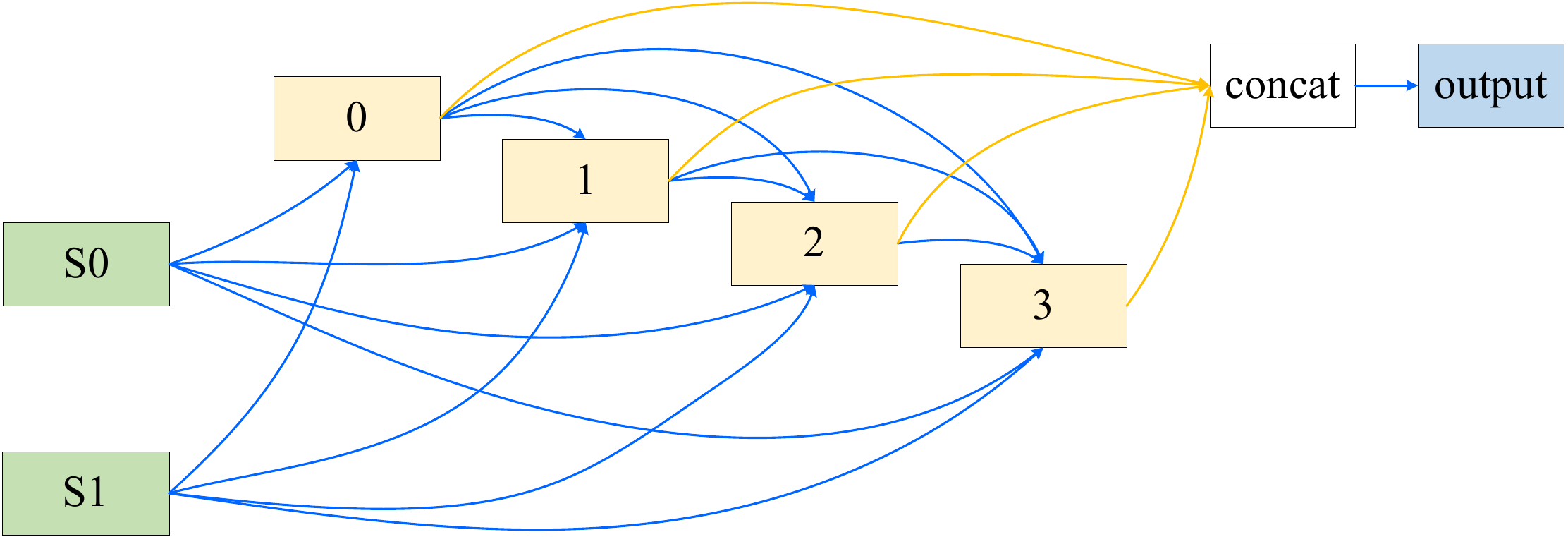}
    \end{center}
    \caption{A dual-bottom cell contains seven nodes. Nodes S0 and S1 are feature maps outputted from the two previous layers.}
    \label{fig:dual}
\end{figure}

\subsection{Basic search units}

\subsubsection{Single-bottom cells}

We search for two kinds of computation cells (normal and reduction), then stack them together to form a convolutional network. The function of the normal cell is to enhance the expressive performance of the model. Reduction cells are used for reducing the size of feature maps to half of the current input and doubling the number of channels. A cell is a directed acyclic graph (DAG) with ordered sequence nodes. As illustrated in Fig.\ref{fig:single}, we propose an architecture with only one input (single-bottom) consisting of 6 nodes in one cell. Input node S represents previous features. Nodes 0, 1, 2, and 3 are intermediate features that are concatenated as the output of the current cell.

\subsubsection{Dual-bottom cells}

The performance of single-bottom architecture may not be good enough because only the most crucial edge is left during inference to simplify cells and reduce calculations.
To preserve more information, the dual-bottom cell is considered to balance performance and computing. The dual-bottom cell is shown in Fig.\ref{fig:dual}. The current input nodes S0 and S1 are outputs from the previous two layers. There are seven nodes for one cell. Search space is the same with single-bottom cells.

\subsection{Operation candidates}

We define eight operation candidates (\aka search space) for each blue edge in Figs.\ref{fig:single} and \ref{fig:dual}: none, 3$\times$3 max pooling, 3$\times$3 average pooling, 3$\times$3 depth-wise separable convolution, 5$\times$5 depth-wise separable convolution, skip connection, 3$\times$3 dilated convolution, 5$\times$5 dilated convolution. The normal and reduction cells are stacked repeatedly to form the final architecture. We keep the most crucial edge in inference to simplify architectures and reduce computations. We use a differential architecture search method to convert all discrete operations into a continuous search space by a softmax function. In this way, the architecture search task is simplified to learn a set of continuous variables,
\begin{equation}
    \bar{o}^{(i,j)}(x)=\sum_{o\in\mathcal{O}}\frac{\exp(\alpha_o^{(i,j)})}{\sum_{o^\prime\in\mathcal{O}}\exp(\alpha_{o^\prime}^{(i,j)})}o(x),
\end{equation}
where $x$ represents feature maps, $\mathcal{O}$ denotes the search space where each operation candidate $o\in\mathcal{O}$ represents convolution or pooling.

\begin{figure*}[t!]
    \begin{center}
        \includegraphics[width=\linewidth]{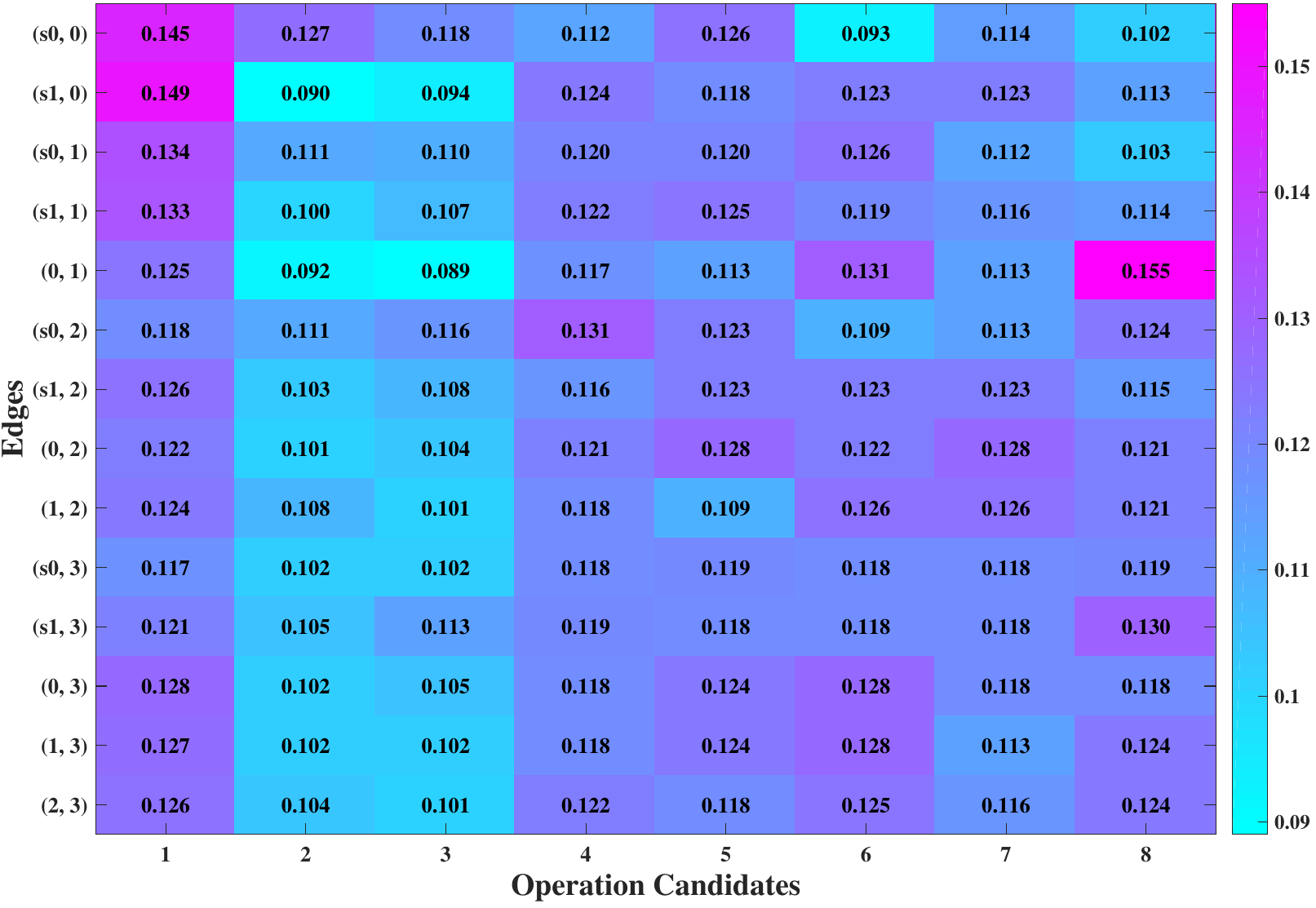}
    \end{center}
    \caption{Architecture parameters $\alpha$ for a dual-bottom cell. Rows and columns represent 14 edges and 8 operation candidates, respectively.}
    \label{fig:var}
\end{figure*}

\begin{figure*}[t!]
    \begin{center}
        \includegraphics[width=\linewidth]{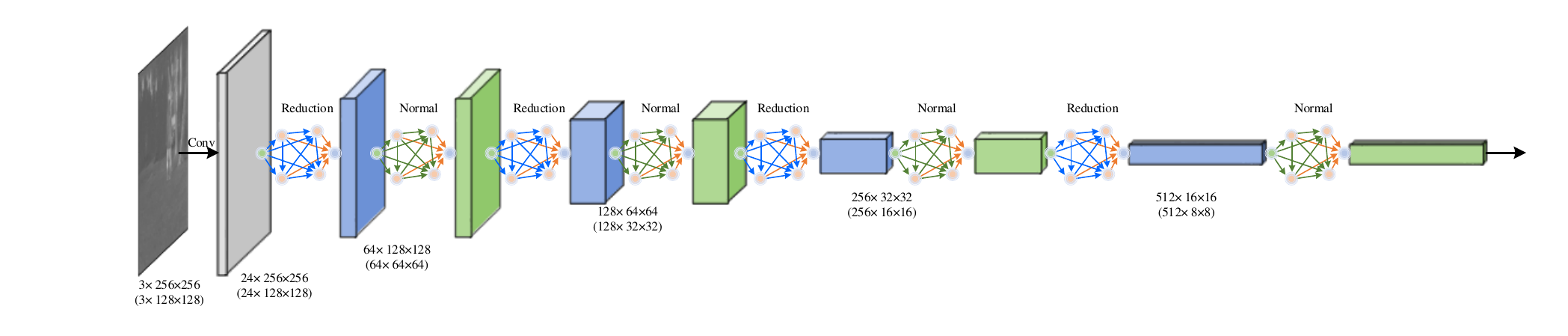}
    \end{center}
    \caption{The overview of network architecture sample with only one input (\aka single-bottom).}
    \label{fig:archsingle}
\end{figure*}

\begin{figure*}[t!]
    \begin{center}
        \includegraphics[width=\linewidth]{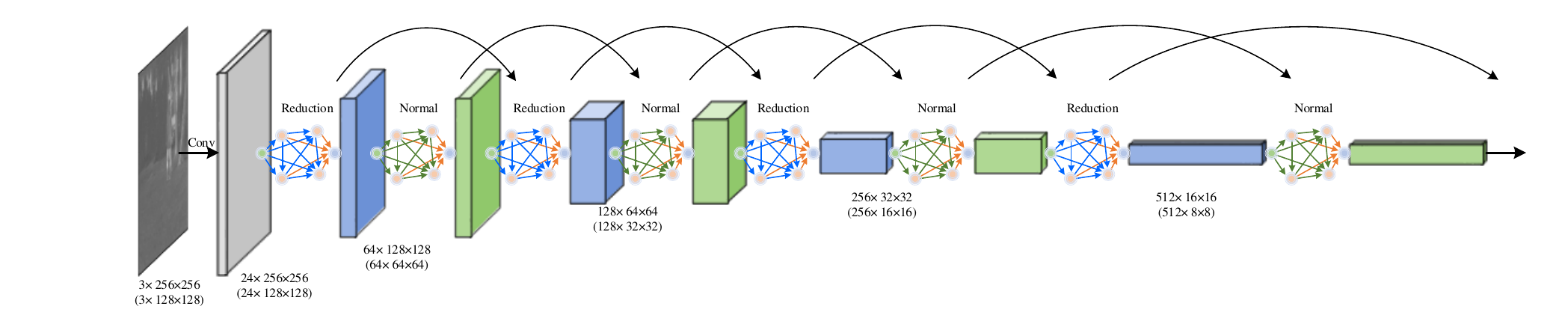}
    \end{center}
    \caption{The overview of network architecture sample with two inputs (\aka dual-bottom).}
    \label{fig:archdual}
\end{figure*}

The continuous variables $a_o^{(i,j)}$ indicates the mixing weight associated with the operation $o$ between nodes $i$ and $j$ and forms a matrix $\alpha$. For Instance, dual-bottom cells are encoded as a matrix with 14 rows (edges) and 8 columns (operation candidates) as shown in Fig.\ref{fig:var}. The matrix represents the topology structure for a cell, named architecture parameters. Other operation parameters are marked as $\omega$. The network architecture samples with single- and dual-bottom cells are illustrated in Figs.\ref{fig:archsingle} and \ref{fig:archdual}, respectively. Size of input image is $C\times H\times W$, specifically $3\times 256\times 256$ and $3\times 128\times 128$ for training and test regions. Blue blocks represent reduction cells, which reduce the size of feature maps and double the number of channels. Green blocks are normal cells utilized to enhance the expressive performance of the network. Searching architecture parameters $\alpha$ and the operation parameters $\omega$ is a bi-level optimization problem, they are thus updated alternately. As for $\alpha$, it needs to optimize two kinds of the matrix where $\alpha_1$ is shared by all the normal cells and $\alpha_2$ is shared by all the reduction cells.

\subsection{Random channel selection}

\begin{figure*}[t!]
    \begin{center}
        \includegraphics[width=\linewidth]{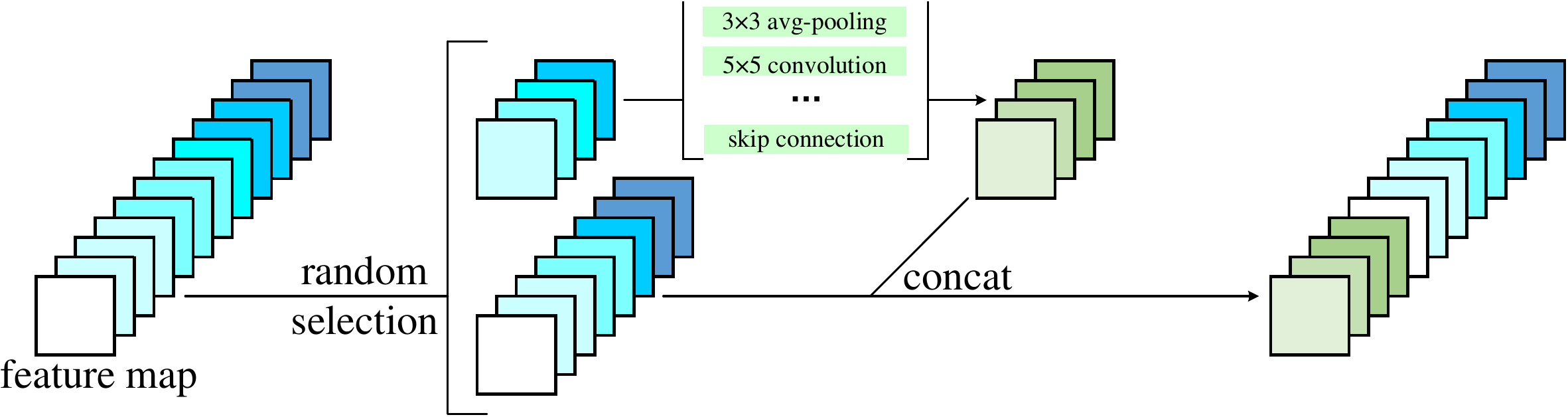}
    \end{center}
    \caption{Illustration of the random channel selection. Instead of calculating all channels, we select a part of channels and feed them into operation candidates, then concatenate with the rest of the unchanged channels.}
    \label{fig:random}
\end{figure*}

Calculating and saving all feature maps during the search process takes a lot of computing resources. The differentiable NAS still suffers from the issue of high GPU memory consumption. In order to solve this problem and speed up the search process with limited resources, we randomly select a subset of channels and then send them into operation candidates while keeping the rest unchanged. The sampling rate of the channel is $1/4$. Finally, we concatenate the features depth-wise, as shown in Fig.\ref{fig:random}. They are calculated as follows,
\begin{equation}
    \bar{o}^{(i,j)}(x;r^{(i,j)})=\sum_{o\in\mathcal{O}}\frac{\exp(\alpha_o^{(i,j)})}{\sum_{o^\prime\in\mathcal{O}}\exp(\alpha_{o^\prime}^{(i,j)})}o(r^{(i,j)}x)+(1-r^{(i,j)})x,
\end{equation}
where $\alpha$ is the searched network architecture, $r^{(i,j)}x$ and $(1-r^{(i,j)})x$ respectively represent selected channels and the rest channels.

\subsection{Joint supervision}

We retrain the searched architecture with joint supervision. To keep the pedestrian identity and separate the foreground from the background, we take a classification loss as the first supervision as,
\begin{equation}
    \mathcal{L}_{CLS}=-\sum_{k}\big[y_k\log(h_k)+(1-y_k)\log(1-h_k)\big],
\end{equation}
where $y_k$ denote the ground-truth label of the $k$-th sample, $h_k$ is denoted as the probability belong to the foreground predicted by the searched network.

TIR-PT is more complex than object recognition and classification because of the great intra-variations in each class and the variety of factors such as thermal crossover and intensity variation. We take triplets into account so that the output feature maps from the searched neural network can be utilized directly. Sample anchor, positive and negative randomly, may render the combinations of triplets to grow cubically as the dataset gets larger for classical triplets. It is impractical to train the model for a long enough time. In order to mine hard triplets, we adopt batch hard triplet loss as a supervision. For each batch, we sample $M$ pedestrian sequences (\aka classes) randomly from the training dataset and then pick $N$ image frames for each class. Therefore, a batch contains $M\times N$ image frames. We calculate the distance between feature maps extract from different image frames, hard positive (pedestrians with remarkably different appearances or scales picked from the same video sequence), and hard negative (pedestrians with similar appearances or scales picked from different video sequences) within the batch. The batch hard triplet loss function is presented as follows.
\begin{equation}
    \mathcal{L}_{BHTri}=\sum_{m=1}^{M}\sum_{n=1}^{N}\big[\delta+\max_{p=1,\ldots,N}\mathcal{D}(f_\alpha(x_m^n),f_\alpha(x_m^p))-\min_{\substack{s=1,\ldots,M\\ t=1,\ldots,N\\ s\neq m}}\mathcal{D}(f_\alpha(x_m^n),f_\alpha(x_s^t))\big]_+,
\end{equation}
where $\delta$ is margin which equals 0.3, $\mathcal{D}$ is a metric function measuring distances in the embedding space, $f_\alpha$ denote distance and feature maps, $x_n^m$ denotes $n$-th image frames of $m$-th pedestrian video sequence, $[u]_+$ represents $\max(u,0)$.

We try to narrow the gap between features extracted from the same pedestrian class with different image frames. We need to make the features of the same class form a single cluster, and the same pedestrian is also required to collapse to a small point eventually. The distances of the intra-class may be further than those of the inter-class under the supervision of triplet loss. In order to minimize the intra-class distances, we utilize center loss to learn a center for features of each class. The function is computed as follows,
\begin{equation}
    \mathcal{L}_{CT}=\sum_{u=1}^\mathcal{B}\mathcal{D}(f_\alpha(x_{h_u}),f_\alpha(x_{y_u})),
\end{equation}
where $\mathcal{B}$ is the batch size, $f_\alpha(x_{h_u})$ and $f_\alpha(x_{y_u})$ represent the feature maps extracted from the predicted and the ground-truth bounding boxes of the $u$-th sample, respectively.

We train the searched architecture under joint supervision of classification loss, batch hard triple loss, and center loss as follows,
\begin{equation}
    \mathcal{L}=\mathcal{L}_{CLS}+\mathcal{L}_{BHTri}+\gamma\mathcal{L}_{CT},
\end{equation}
where $\gamma$ is used for balancing the center loss and set to 0.5.

\section{Experiments}\label{sec:4}

\begin{figure}[t!]
    \begin{center}
        \includegraphics[width=\linewidth]{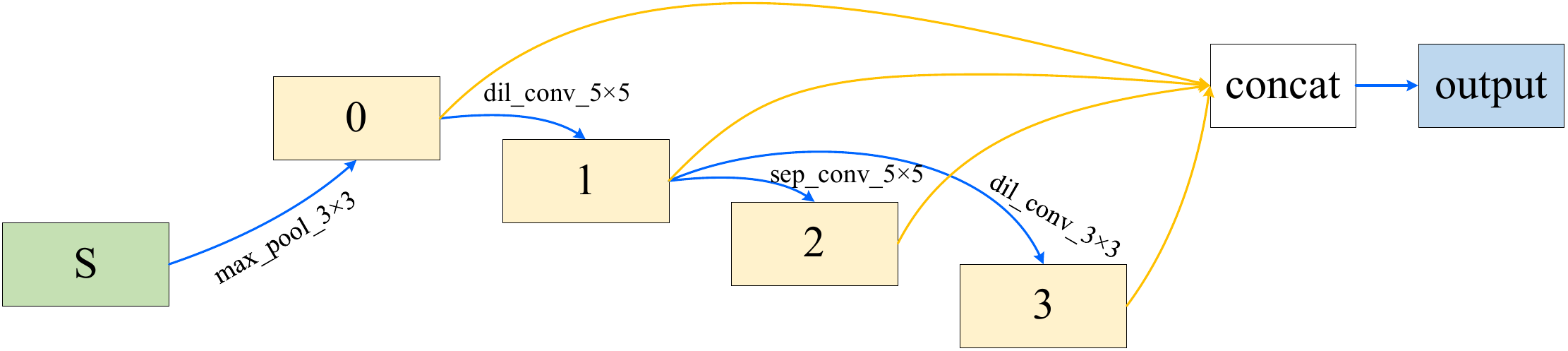}
    \end{center}
    \caption{Illustration of the searched normal single-bottom cell. In order to simplify architecture and reduce calculations, each intermediate node takes the most important edge as input. It is similar to attention-based pruning.}
    \label{fig:singlen}
\end{figure}

\begin{figure}[t!]
    \begin{center}
        \includegraphics[width=\linewidth]{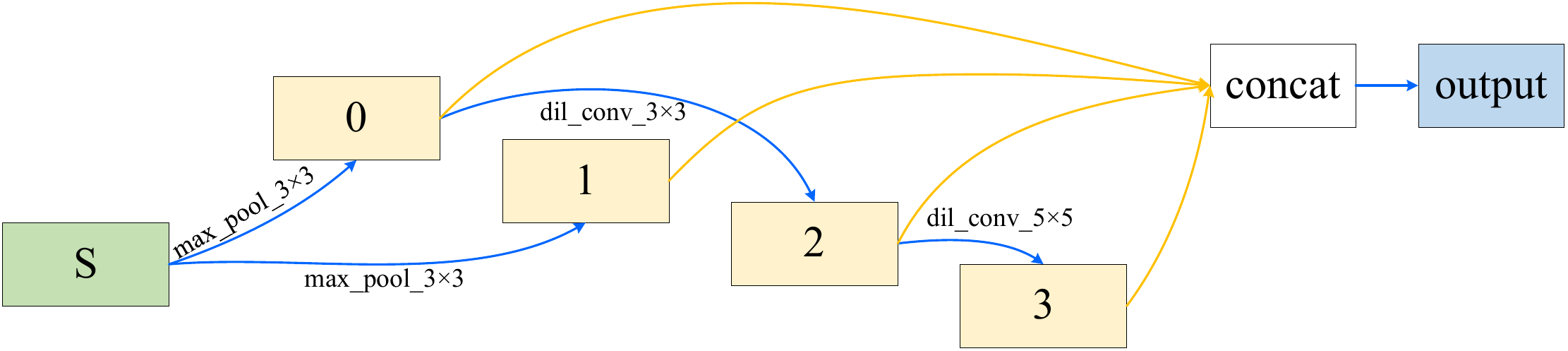}
    \end{center}
    \caption{Illustration of the searched reduction single-bottom cell. The function of reduction cell is to reduce the size of feature maps and double channels. Intermediate layer only takes the most important edges as input. The last output is depth-wise concatenation of four intermediate nodes.}
    \label{fig:singler}
\end{figure}

\subsection{Datasets and metrics}

Two popular TIR-PT datasets are used in our experiments: LSOTB-TIR~\cite{lsotb} and PTB-TIR~\cite{ptb}. LSOTB-TIR includes 1,280 TIR video sequences for evaluation and 120 for training, the number of total image frames is more than 600K. PTB-TIR consists of 60 TIR pedestrian video sequences with a total of 30K image frames.

LSOTB-TIR uses precision, success, and normalized precision as evaluation metrics. Precision is the ratio of frames with center location error between the predicted bounding box and the ground truth below a specified threshold (20 pixels) relative to the size of the video sequence. Normalized precision adjusts precision by considering the scale of the pedestrian or the size of the image frame. Success evaluates the overlap between the predicted bounding box and the ground truth, usually represented as the area under the curve (AUC) in a success plot. PTB-TIR used the same precision and success as LSOTB-TIR to report the tracking performance.

\subsection{Implementation details}

In the searching process, random erasing, normalization, random horizontal flipping with 0.5, and padding the resized image 10 pixels with zero values are applied as the data augmentation. We hold out half of the training data to update architecture parameters $\alpha$. The other half of the training data can be viewed as a search validation set used to update operation parameters $\omega$. We utilize Adam~\cite{kingma2014adam} as the optimizer for $\alpha$, with an initial learning rate of 0.02, decay rates $\beta_1=0.5$ and $\beta_2=0.999$. We use stochastic gradient descent (SGD) with momentum to optimize the weights $\omega$. The initial learning rate starts from 0.1 and anneals down to 0.001 following a cosine schedule without restart for each epoch~\cite{loshchilov2016sgdr}. We search architecture for 200 epochs. In the retraining process, we apply horizontal flip and normalization as data augmentation. SGD with momentum is deployed as the optimizer with an initial learning rate of 0.1 decayed with a cosine annealing. Each batch contains 8 pedestrians (\aka $M$), and each pedestrian has 8 image frames (\aka $N$), resulting in a batch size of 64.

The training dataset of LSOTB-TIR~\cite{lsotb} is utilized to search and retrain the network architecture, we resize each training and test regions to 128$\times$128 and 256$\times$256. Except for the aforementioned details, the rest of the hyperparameters and configurations are are the same as DiMP~\cite{dimp}.
We implement our method in Python using PyTorch with a cloud server with an Intel$^\circledR$ Xeon$^\circledR$ Gold 6148v4 CPU @ 2.2 GHz CPU with 256 GB RAM and a NVIDIA$^\circledR$ Tesla$^\circledR$ V100-SXM2 GPU with 16 GB VRAM.

\subsection{Ablation studies}

\begin{figure}[t!]
    \begin{center}
        \includegraphics[width=\linewidth]{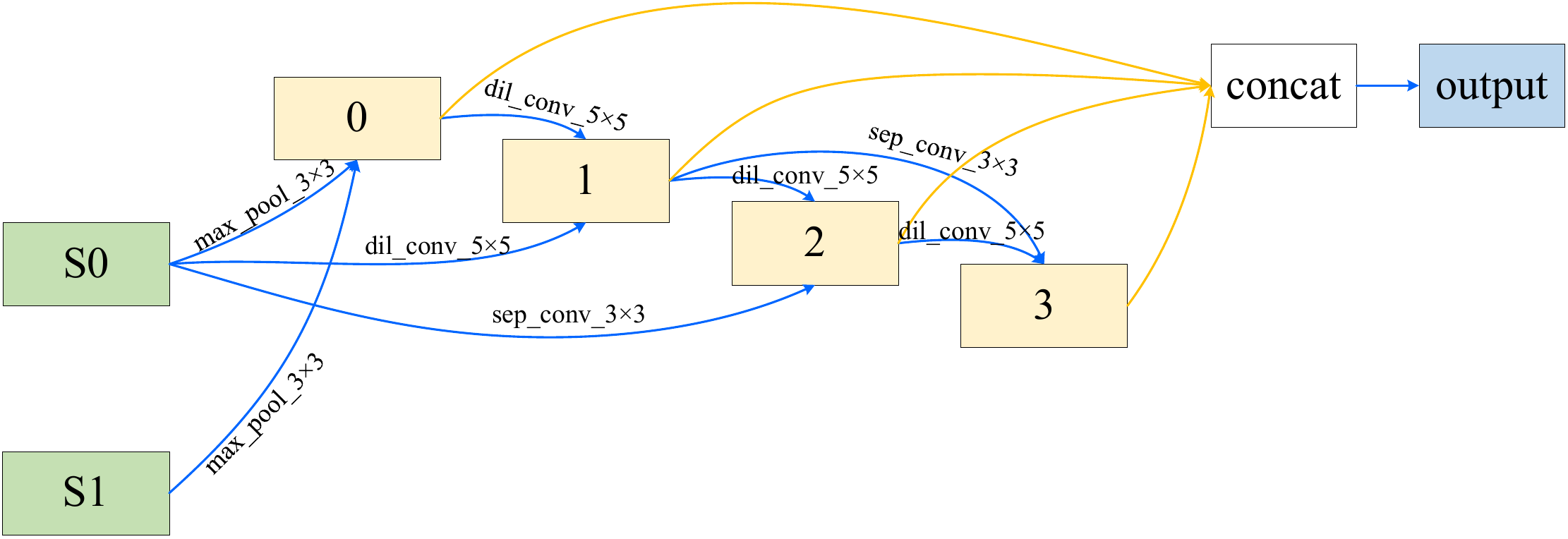}
    \end{center}
    \caption{Illustration of the searched normal dual-bottom cell. To balance efficiency and parameters, each intermediate layer has two important inputs according architecture matrix.}
    \label{fig:dualn}
\end{figure}

\begin{figure}[t!]
    \begin{center}
        \includegraphics[width=\linewidth]{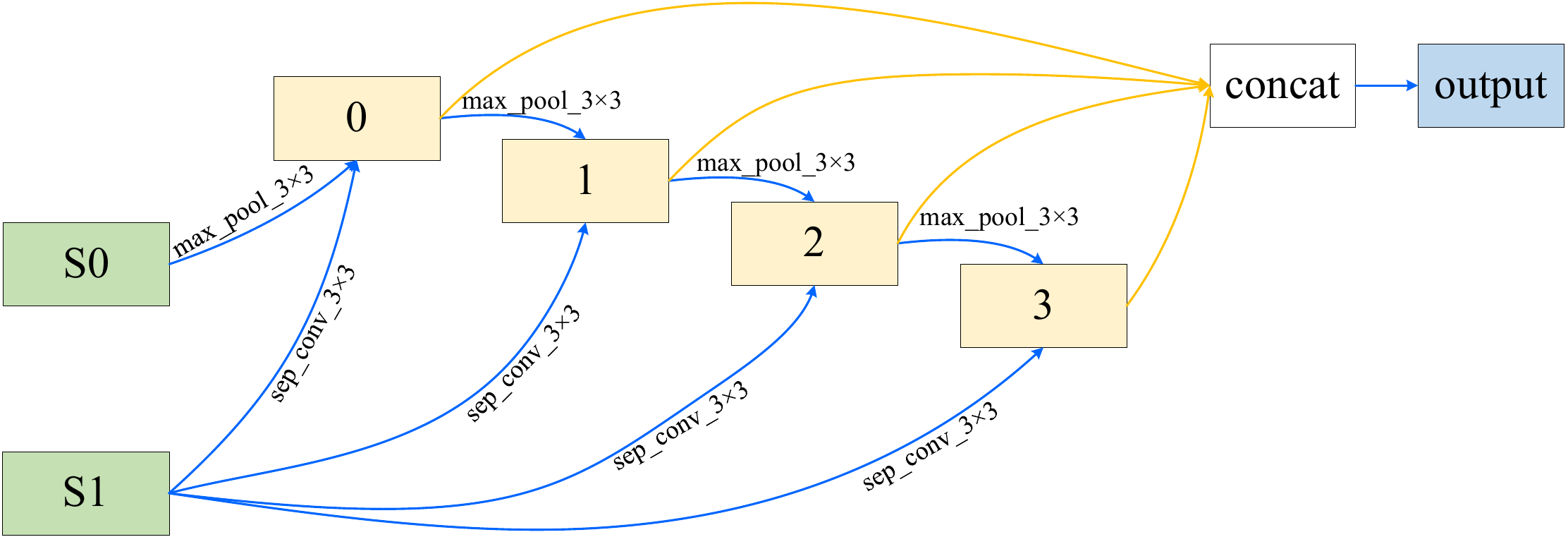}
    \end{center}
    \caption{Illustration of the searched reduction dual-bottom cell. It takes two important layers as current input according to architecture parameters.}
    \label{fig:dualr}
\end{figure}

\begin{figure}[t!]
    \begin{center}
        \includegraphics[width=\linewidth]{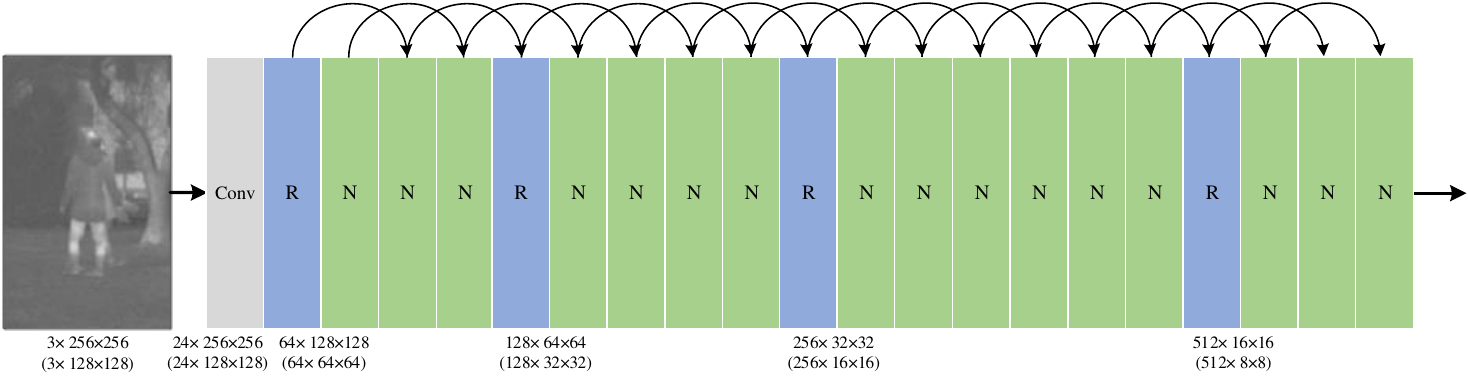}
    \end{center}
    \caption{Illustration of the final searched network architecture. ``R'' and ``N'' indicate the searched reduction and normal dual-bottom cells, respectively.}
    \label{fig:archfinal}
\end{figure}

\begin{table*}[t!]
\centering
\caption{Ablation studies of the proposed method on the LSOTB-TIR benchmark dataset. \#Dims and \#Cells represent the dimensionality of the output feature map and the number of stacked cells of the network architecture, respectively. Single, Dual, Channel, $\mathcal{L}_{CLS}$, $\mathcal{L}_{BHTri}$, and $\mathcal{L}_{CT}$ denote the single-bottom cell, dual-bottom cell, random channel selection, classification loss, batch triplet loss, and center loss, respectively. The best two results are shown in \textcolor{red}{red} and \textcolor{blue}{blue} fonts.}
\label{tab:abla}
\resizebox{\textwidth}{!}{%
\begin{tabular}{c|c|cccccc|ccc|cc|c}
\toprule
\multirow{2}{*}{\#Cells} &
\multirow{2}{*}{\#Dims} &
\multicolumn{6}{c|}{Configurations} &
  \multirow{2}{*}{Precision $\uparrow$} &
  \multirow{2}{*}{Success $\uparrow$} &
  \multirow{2}{*}{Norm. Precision $\uparrow$} &
  \multirow{2}{*}{Params $\downarrow$} &
  \multirow{2}{*}{FLOPs $\downarrow$} &
  \multirow{2}{*}{GPU days $\downarrow$} \\\cline{3-8}
& & Single & Dual & Channel & $\mathcal{L}_{CLS}$ & $\mathcal{L}_{BHTri}$ & $\mathcal{L}_{CT}$ &  &  &  &      &      &                      \\
\midrule
\multirow{6}{*}{16} & \multirow{5}{*}{512} & \checkmark   &              &              & \checkmark  &             &              & 0.768 & 0.640 & 0.697 & \textcolor{blue}{2.96} M & \textcolor{blue}{0.78} G & \multirow{2}{*}{3.3} \\
& & \checkmark   & \checkmark   &              & \checkmark  &             &              & 0.783 & 0.656 & 0.715 & 4.73 M & 1.35 G &                      \\\cline{3-14}
& & \checkmark   & \checkmark   & \checkmark   & \checkmark  &             &              & 0.785 & 0.656 & 0.715 & \multirow{3}{*}{5.31 M} & \multirow{3}{*}{1.58 G} & \multirow{3}{*}{1.9} \\
& & \checkmark   & \checkmark   & \checkmark   & \checkmark  & \checkmark  &              & 0.792 & 0.665 & 0.717 &      &      &                      \\
& & \checkmark   & \checkmark   & \checkmark   & \checkmark  & \checkmark  & \checkmark   & \textcolor{blue}{0.805} & \textcolor{blue}{0.669} & \textcolor{blue}{0.720} &      &      &\\\cline{2-14}
 & 1024 & \checkmark   & \checkmark   & \checkmark   & \checkmark  & \checkmark  & \checkmark   & \textcolor{red}{0.825} & \textcolor{red}{0.697} & \textcolor{red}{0.733} & 23.43 M  & 2.67 G & 3.1 \\\hline
\multirow{2}{*}{8} & 512 & \checkmark   & \checkmark   & \checkmark   & \checkmark  & \checkmark  & \checkmark   & 0.772 & 0.646 & 0.708 & \textcolor{red}{1.28} M  & \textcolor{red}{0.39} G & \textcolor{red}{0.4} \\\cline{2-14}
 & 1024 & \checkmark   & \checkmark   & \checkmark   & \checkmark  & \checkmark  & \checkmark   & 0.797 & 0.665 & 0.717 & 3.18 M  & 1.06 G & \textcolor{blue}{0.9} \\
\bottomrule
\end{tabular}%
}
\end{table*}

\begin{table*}[t!]
\centering
\caption{Comparison of the state-of-the-art methods on the LSOTB-TIR and PTB-TIR benchmark datasets. The best two results are shown in \textcolor{red}{red} and \textcolor{blue}{blue} fonts.}
\label{tab:comp}
\begin{tabular}{l|ccc|cc}
\toprule
\multirow{2}{*}{Methods} & \multicolumn{3}{c|}{LSOTB-TIR}         & \multicolumn{2}{c}{PTB-TIR} \\\cline{2-6}
                          & Precision $\uparrow$ & Success $\uparrow$ & Norm. Precision $\uparrow$ & Precision $\uparrow$     & Success $\uparrow$     \\
\midrule
DBF~\cite{dbf}                        & \textcolor{blue}{0.770}     & \textcolor{blue}{0.625}   & \textcolor{blue}{0.703}           & \textcolor{red}{0.839}         & \textcolor{blue}{0.626}       \\
DiMP~\cite{dimp}                      & 0.739     & 0.621   & 0.668           & 0.737         & 0.589       \\
ECO-deep~\cite{eco}                   & 0.739     & 0.609   & 0.670           & 0.838         & 0.633       \\
ECO-stir~\cite{ecostir}               & 0.750     & 0.616   & 0.672           & 0.830         & 0.617       \\
HSSNet~\cite{hssnet}                  & 0.515     & 0.409   & 0.488           & 0.689         & 0.468       \\
MCFTS~\cite{mcfts}                    & 0.635     & 0.479   & 0.546           & 0.690         & 0.492       \\
MDNet~\cite{mdnet}                    & 0.750     & 0.601   & 0.686           & 0.817         & 0.593       \\
MLSSNet~\cite{mlssnet}                & 0.596     & 0.459   & 0.549           & 0.741         & 0.539       \\
MMNet~\cite{mmnet}                    & 0.582     & 0.476   & 0.539           & 0.783         & 0.557       \\
SiamFC~\cite{siamfc}                  & 0.651     & 0.517   & 0.587           & 0.623         & 0.480       \\
SiamTri~\cite{siamtri}                & 0.649     & 0.513   & 0.583           & 0.608         & 0.459       \\
TADT~\cite{tadt}                      & 0.710     & 0.587   & 0.635           & 0.740         & 0.560       \\\hline
Ours                                  & \textcolor{red}{0.805}     & \textcolor{red}{0.669}   & \textcolor{red}{0.720}           & \textcolor{blue}{0.832}         & \textcolor{red}{0.643}       \\
\bottomrule
\end{tabular}%
\end{table*}

\begin{figure*}[t!]
\begin{center}
    \subfigure{\includegraphics[width=.32\linewidth]{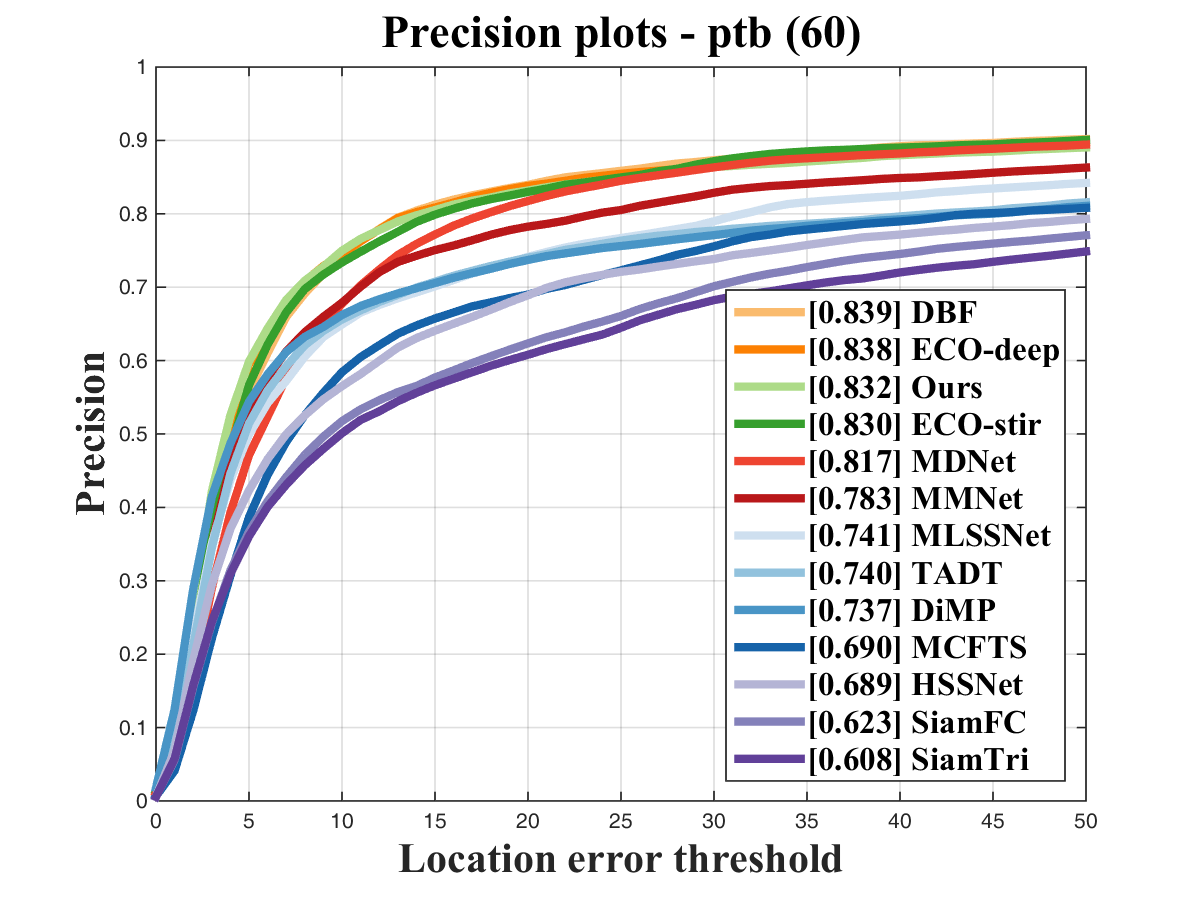}}
    \hspace{0.05em}
    \subfigure{\includegraphics[width=.32\linewidth]{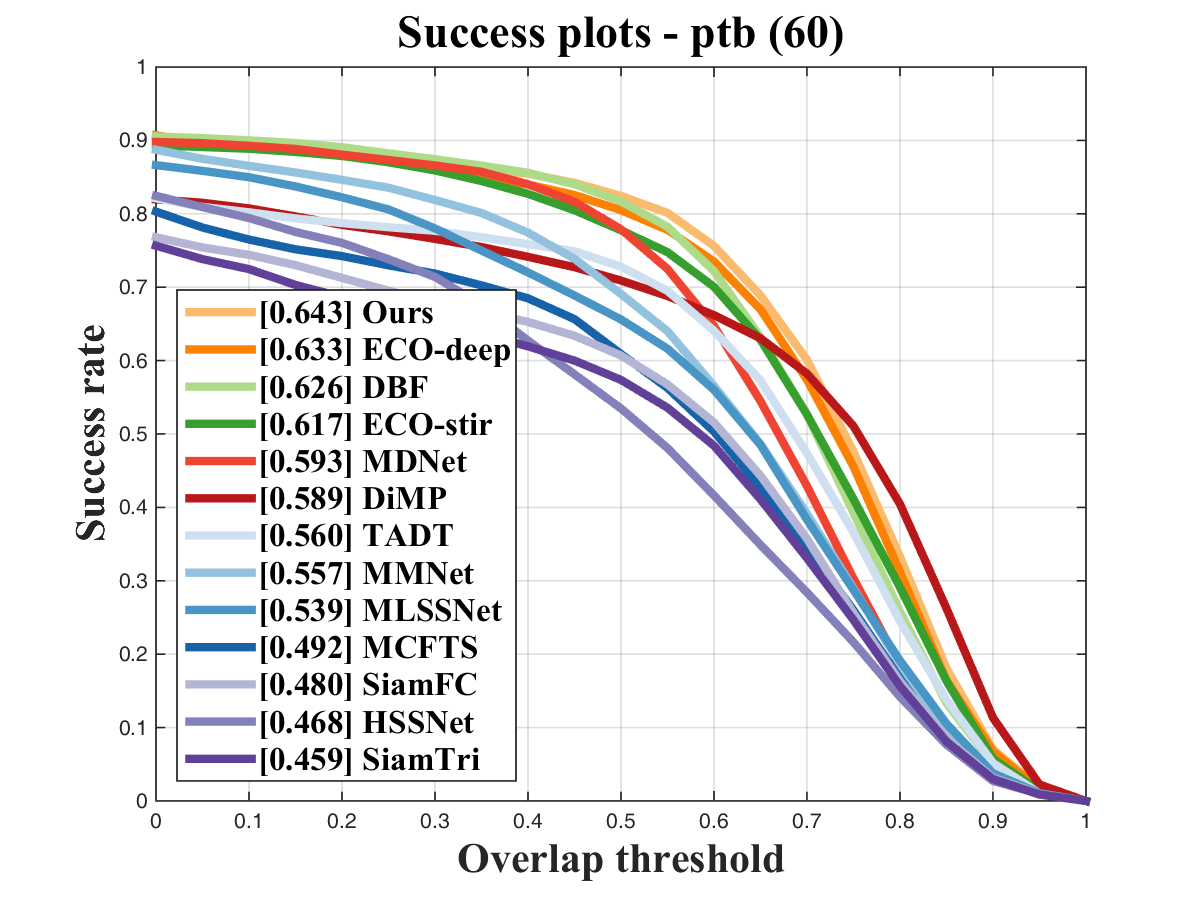}}
    \hspace{0.05em}
    \subfigure{\includegraphics[width=.32\linewidth]{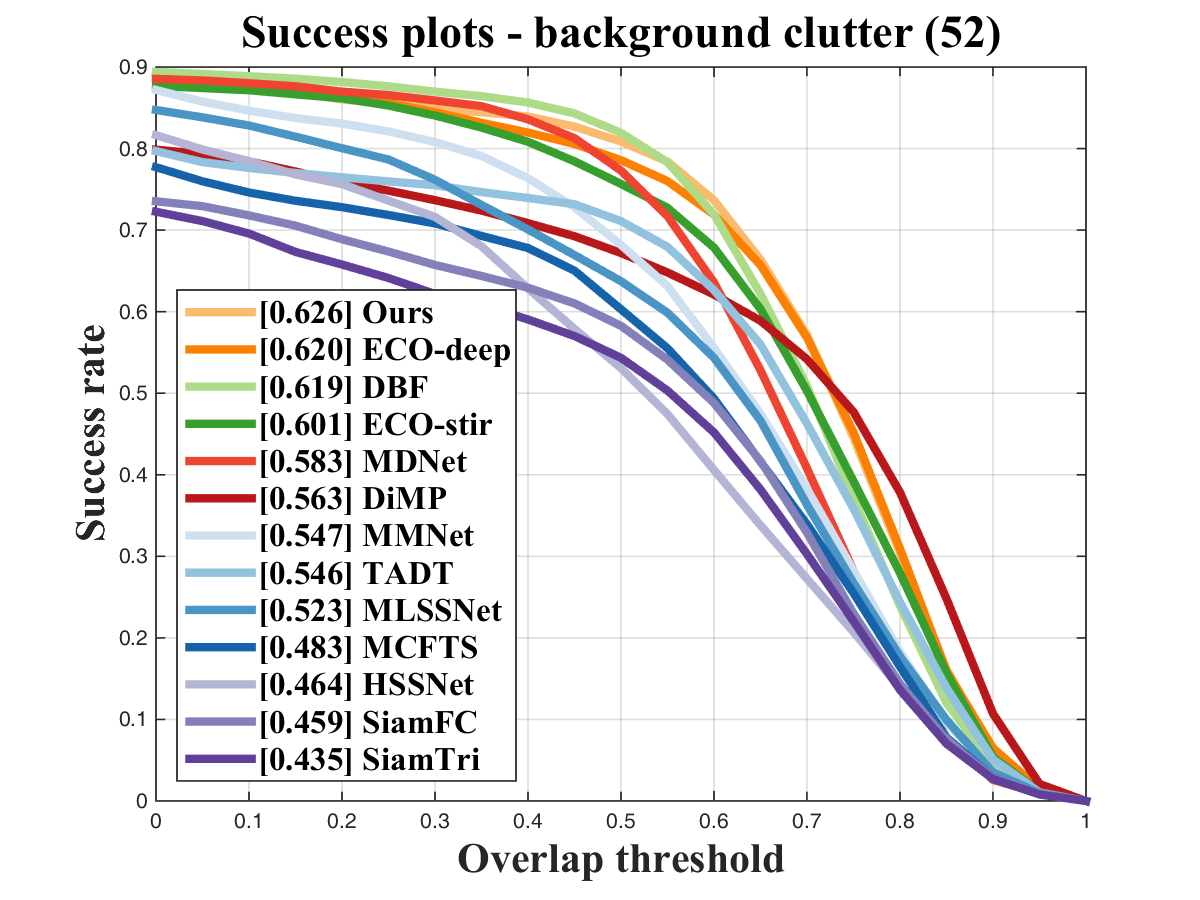}}
    \hspace{0.05em}
    \vfill
    \subfigure{\includegraphics[width=.32\linewidth]{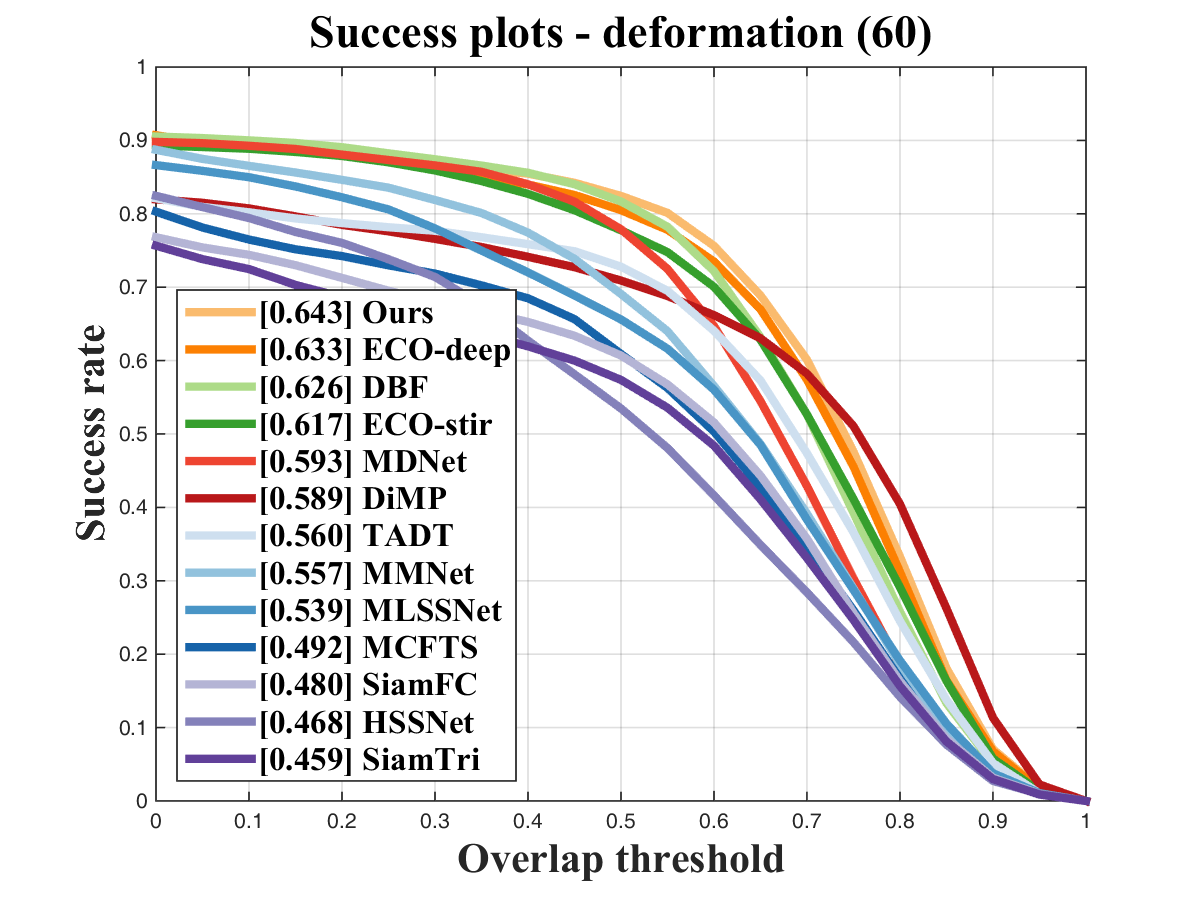}}
    \hspace{0.05em}
    \subfigure{\includegraphics[width=.32\linewidth]{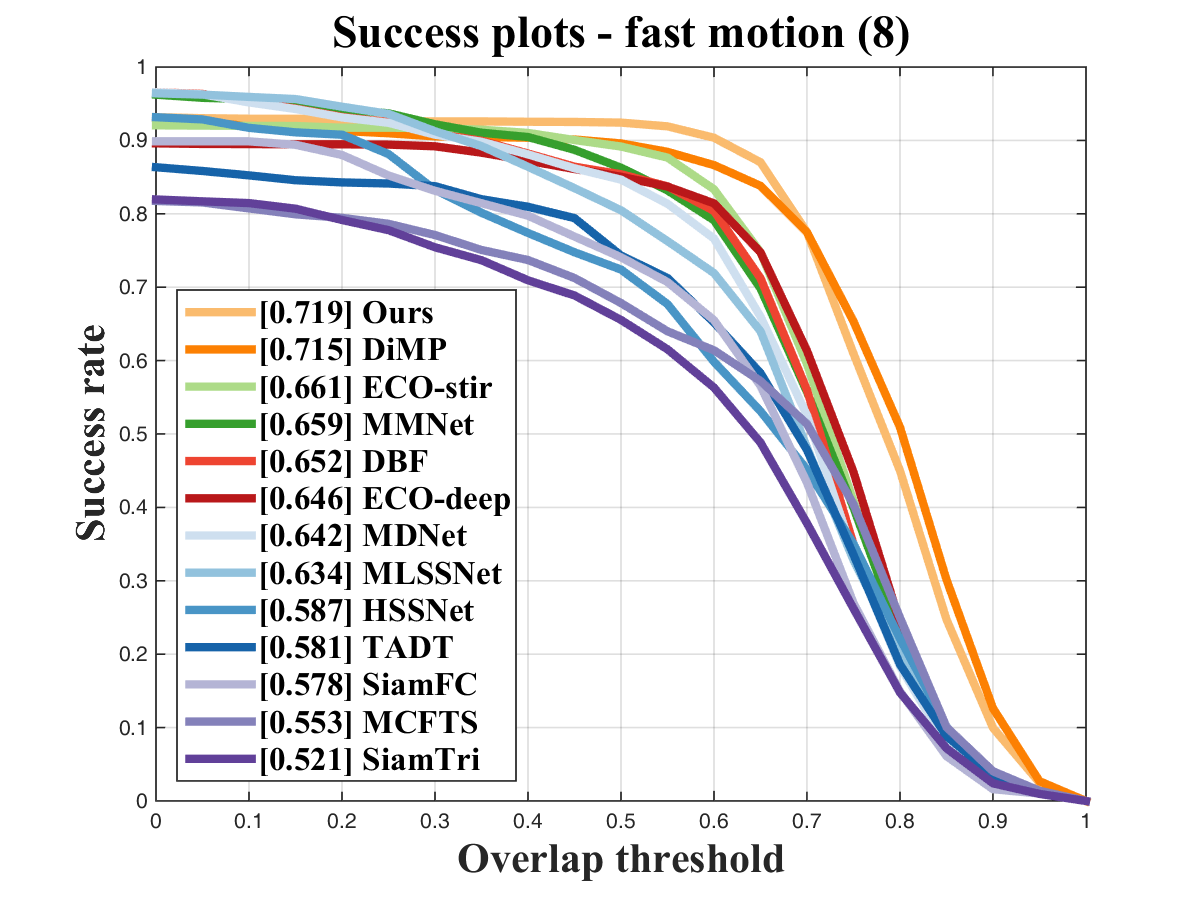}}
    \hspace{0.05em}
    \subfigure{\includegraphics[width=.32\linewidth]{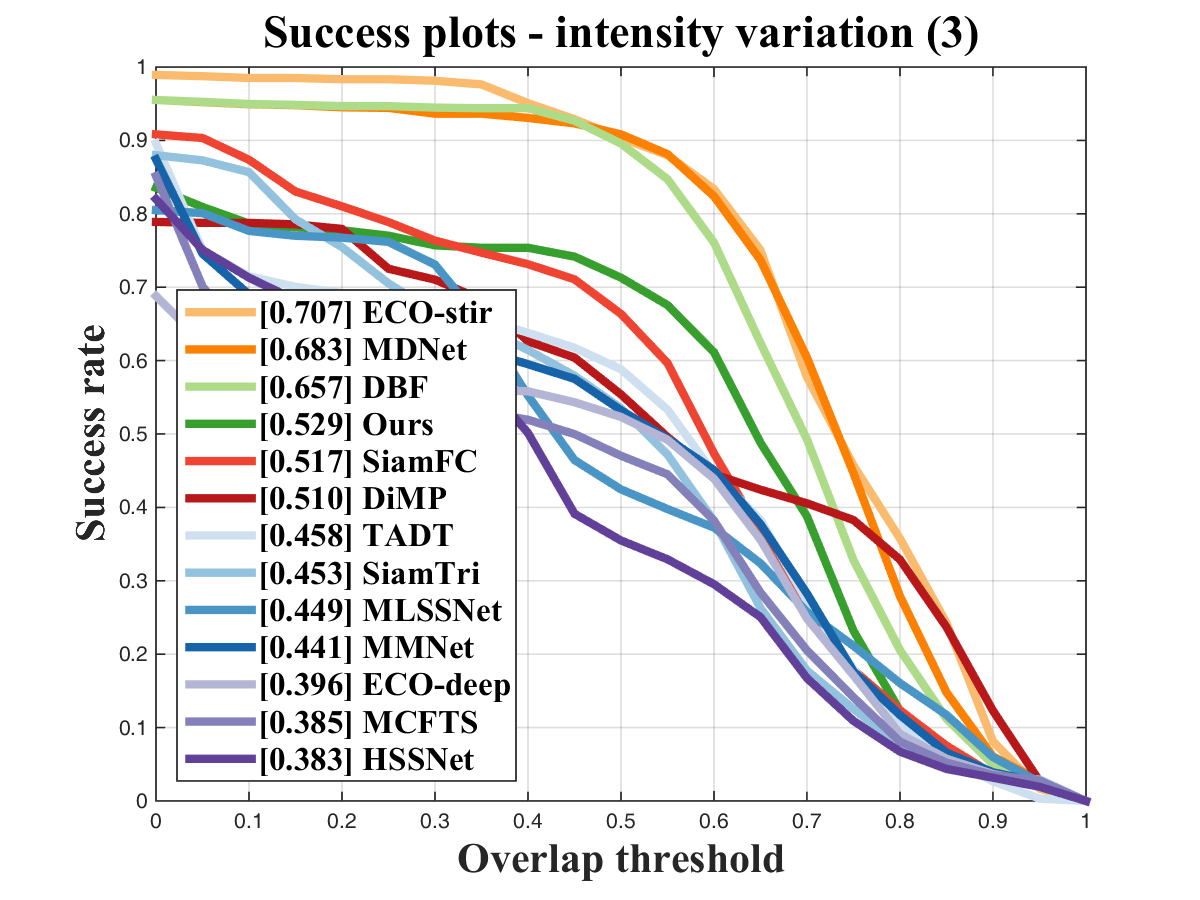}}
    \hspace{0.05em}
    \vfill
    \subfigure{\includegraphics[width=.32\linewidth]{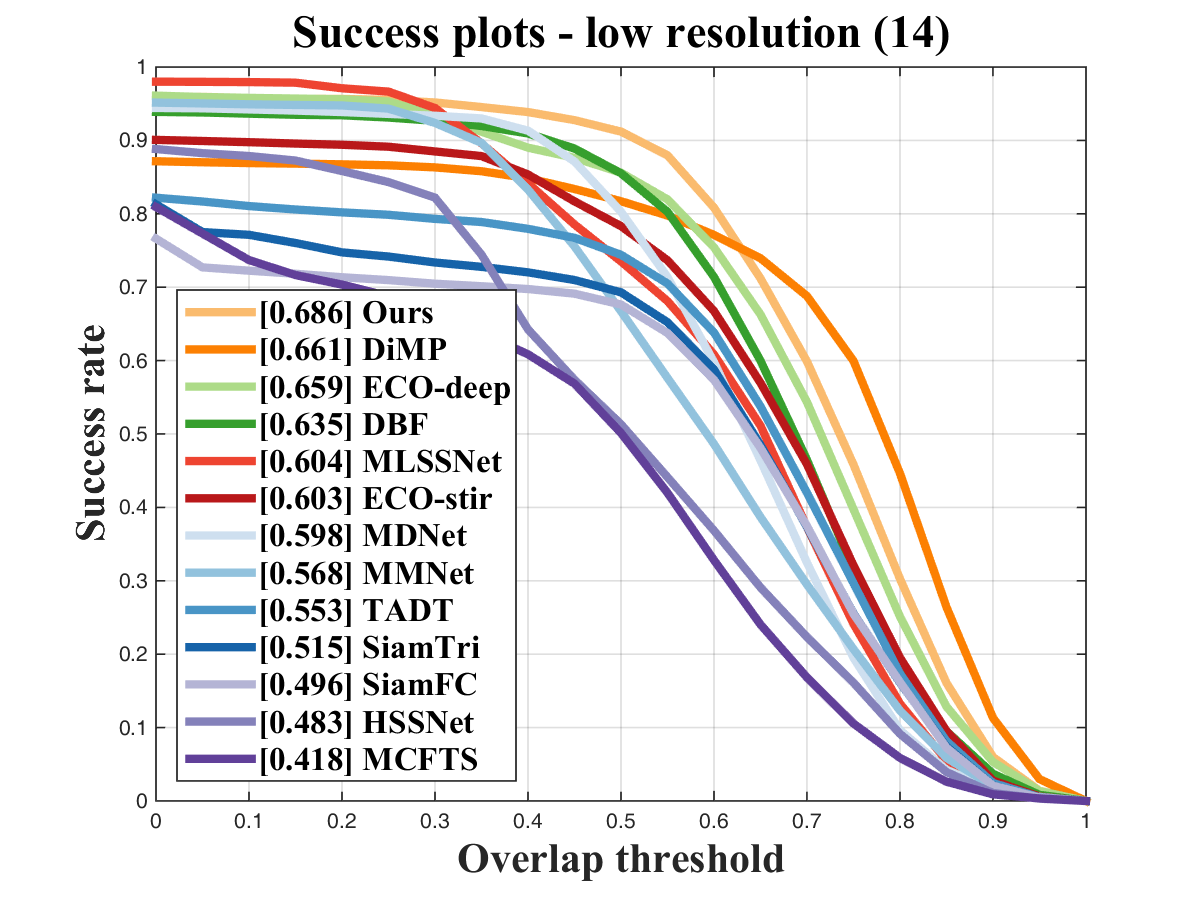}}
    \hspace{0.05em}
    \subfigure{\includegraphics[width=.32\linewidth]{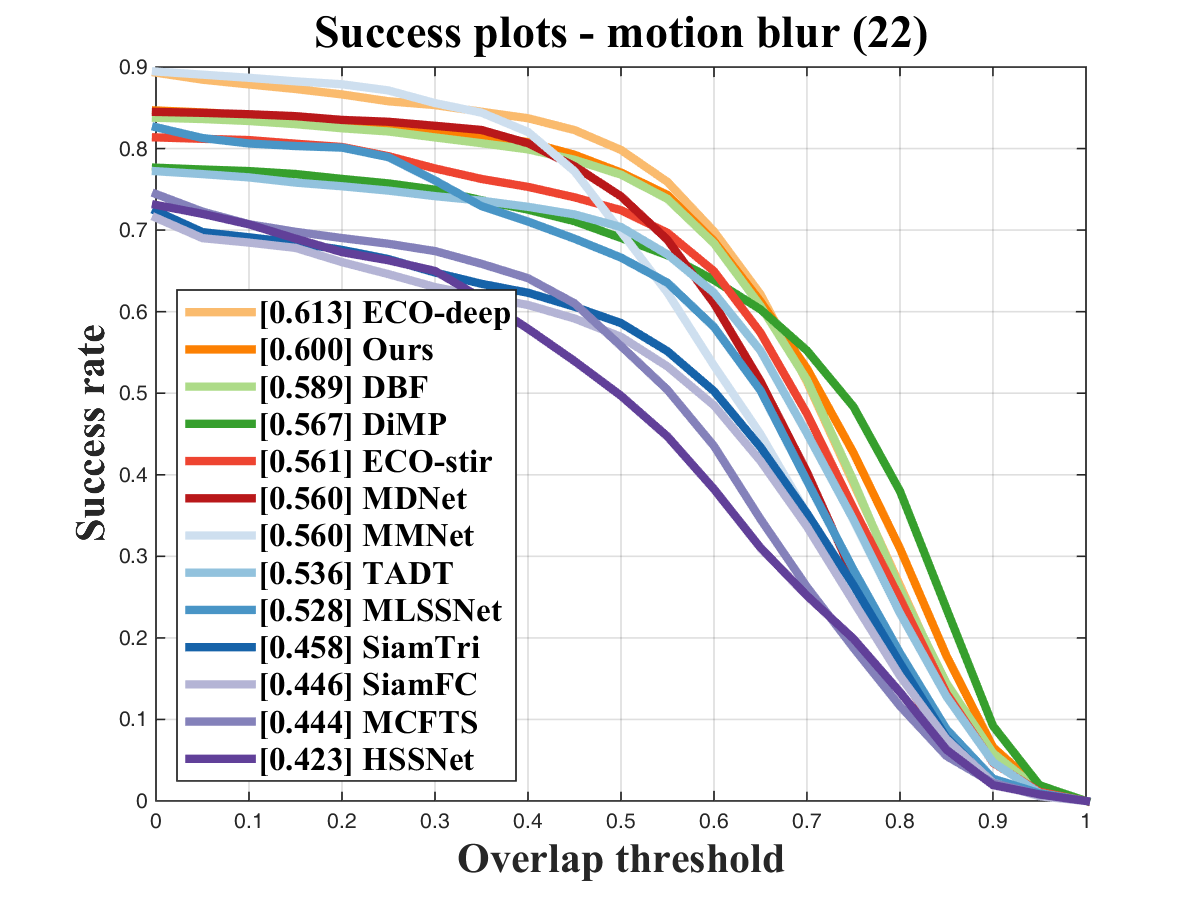}}
    \hspace{0.05em}
    \subfigure{\includegraphics[width=.32\linewidth]{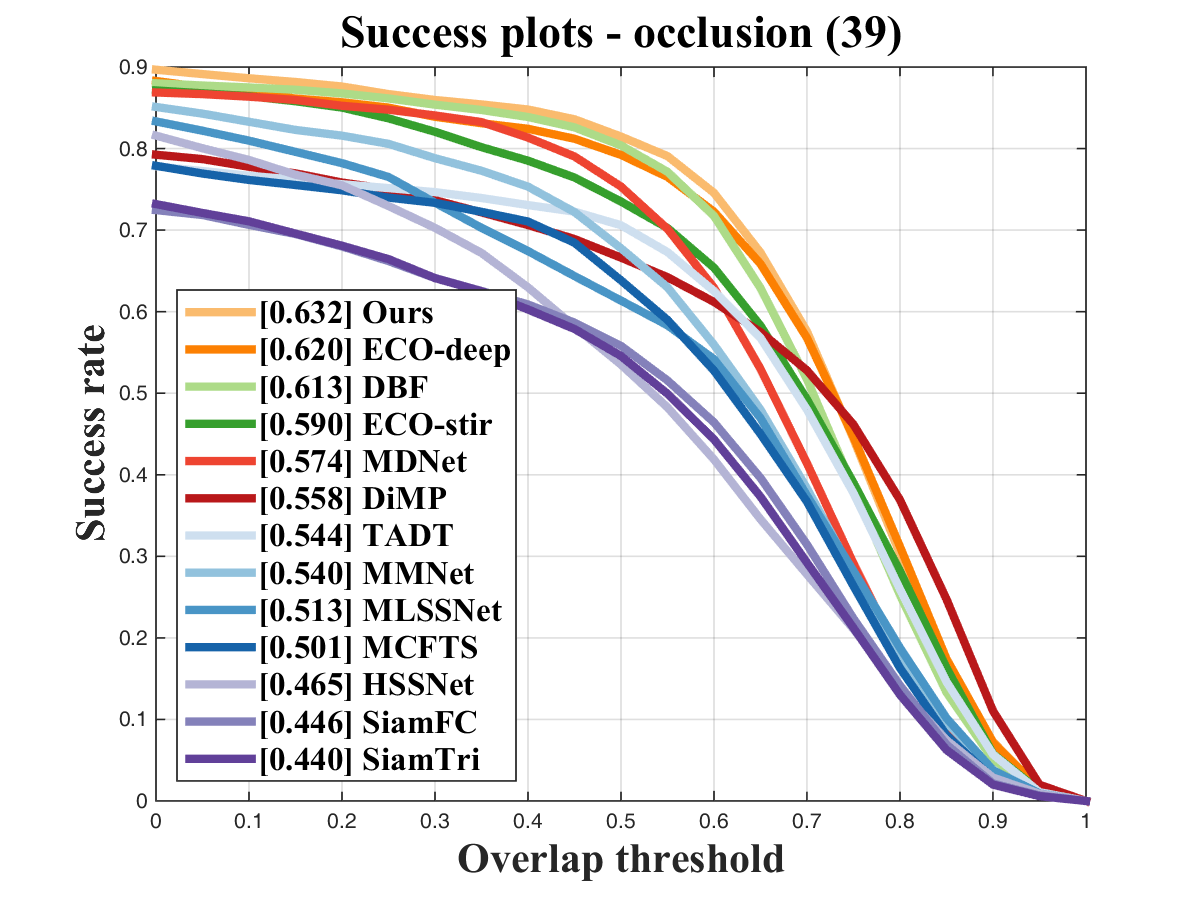}}
    \hspace{0.05em}
    \vfill
    \subfigure{\includegraphics[width=.32\linewidth]{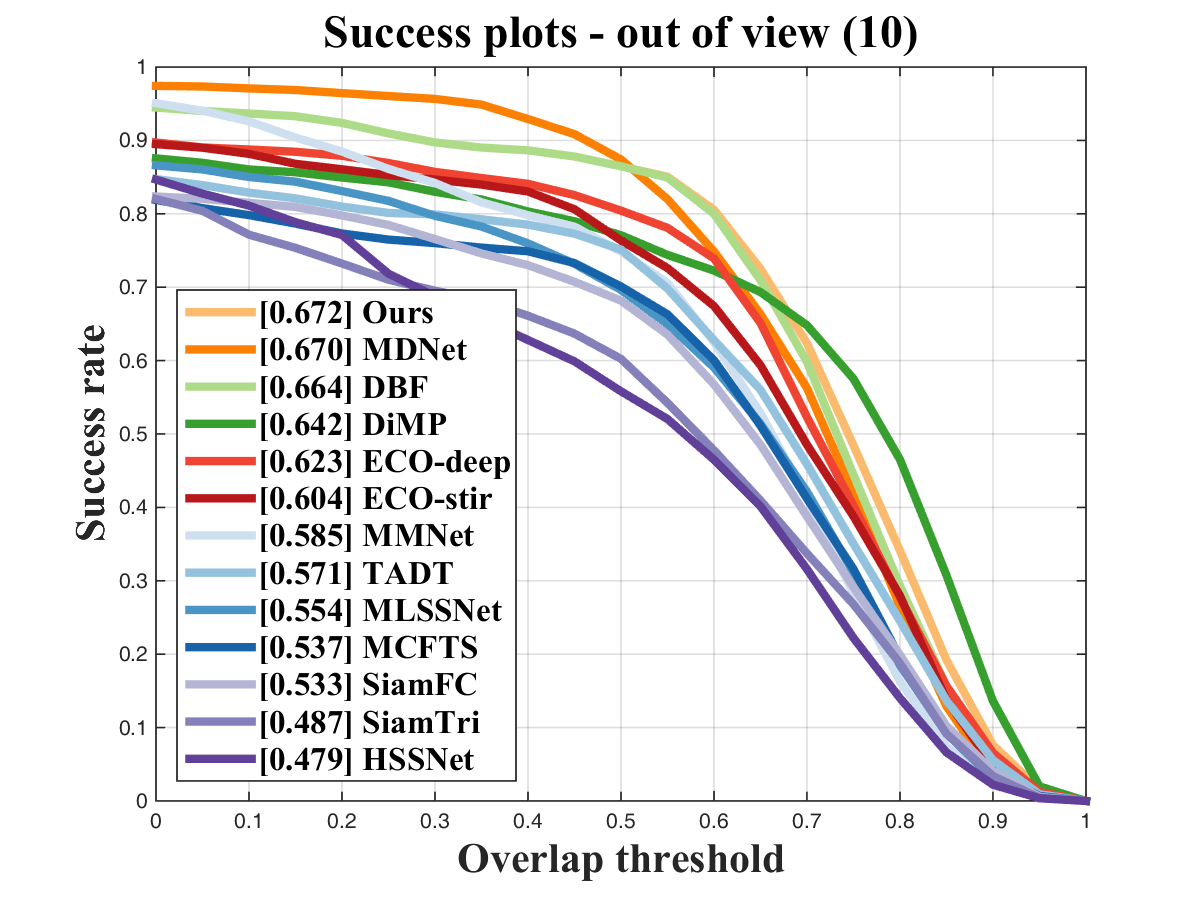}}
    \hspace{0.05em}
    \subfigure{\includegraphics[width=.32\linewidth]{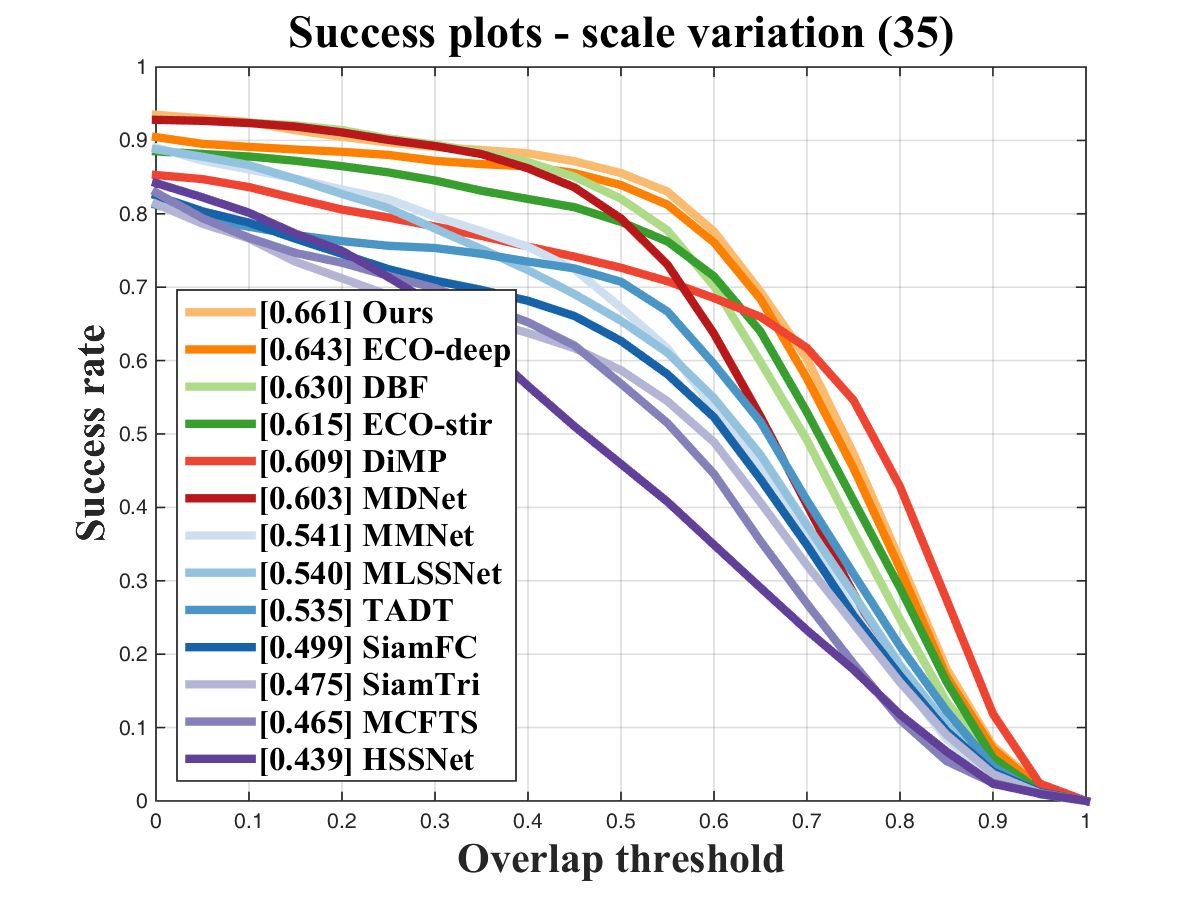}}
    \hspace{0.05em}
    \subfigure{\includegraphics[width=.32\linewidth]{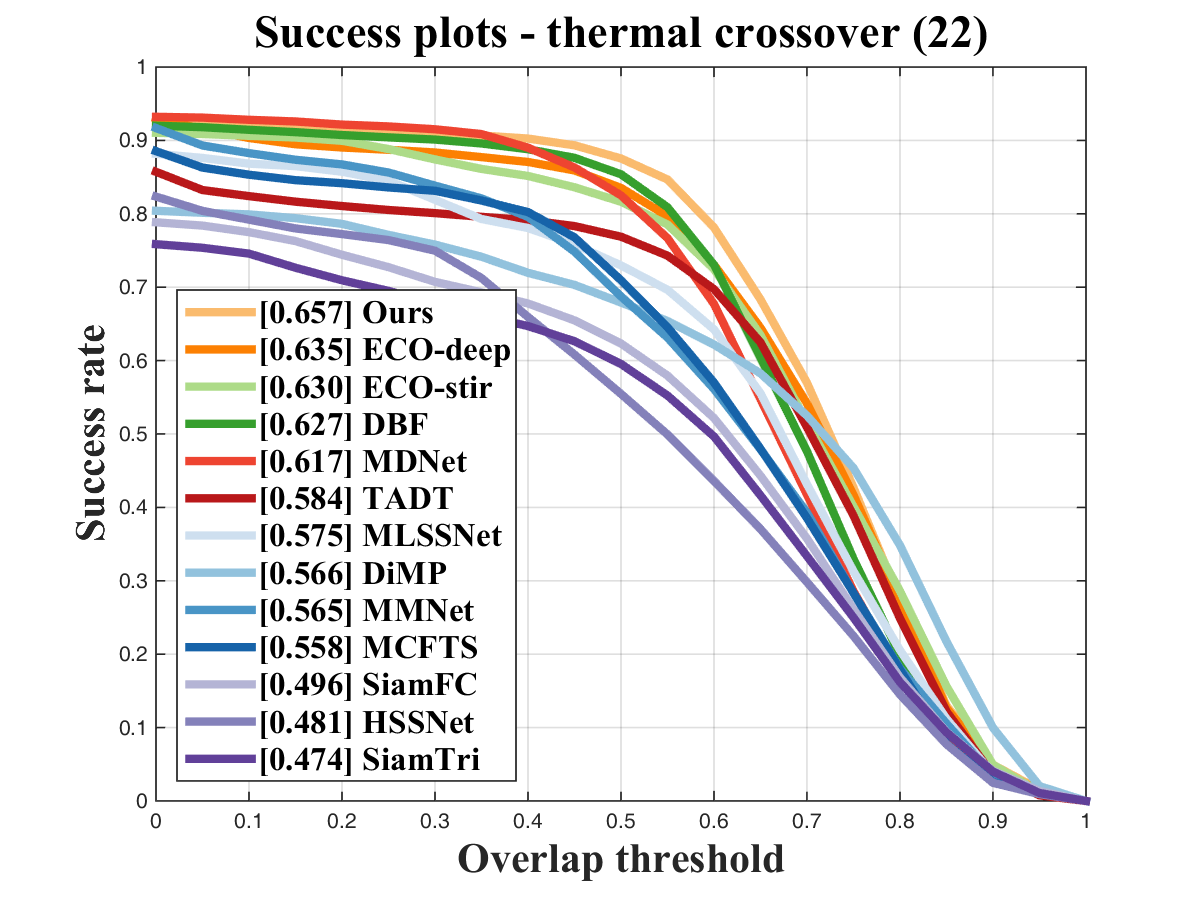}}
\end{center}
\caption{Comparison of the state-of-the-art methods on PTB-TIR challenge subsets with several attributes.}
\label{fig:lsotb}
\end{figure*}

\subsubsection{Effects of single-bottom and dual-bottom cells}

The searched normal single-bottom cell is illustrated in Fig.\ref{fig:singlen}. To simplify architecture, only the most crucial edge is considered as input for the next layer. For instance,  instead of feeding nodes S, 0, 1, and 2 into node 3, we only feed node 2 into node 3 through a 5$\times$5 dilated convolution because the architecture parameters between nodes 2 and 3 are the largest. It’s similar to attention-based pruning. The same principle applies to the reduction single-bottom cell, as shown in Fig.\ref{fig:singler}.
The searched normal and reduction dual-bottom cells are shown in Figs.\ref{fig:dualn} and \ref{fig:dualr}, respectively, which take two nodes as the input. The dual-bottom cells are more complicated than single-bottom cells. We retrain the searched architecture (using random initialization) to evaluate the performance of the network. The results are shown in Table \ref{tab:abla}. The single-bottom cells are very simple and have only one input, thus losing a lot of information from previous layers. Therefore, the parameters and FLOPs are small, while precision and success rates are not good enough. As for the dual-bottom cells, they balance the efficiency and parameters, thus significantly improving the tracking performance.

\subsubsection{Effects of depth and width}

We study the effects of architecture scaling (depth and width) by setting different configurations during the searching and training process. Results are illustrated in Table \ref{tab:abla}. Considering the efficiency, memory, and performance, we use 512 dimensions and 16 cells during searching and training. The setting allows experiments to run on a single GPU.

\subsubsection{Effect of selecting channel randomly}

We randomly select some channels and feed them into 8 operation candidates to reduce memory consumption and speed up the search process. The retrain process follows the setting of calculating all channels. We stack 16 cells (12 normal and 4 reduction cells) to form the final dual-bottom architecture, as shown in Fig.\ref{fig:archfinal}. It takes 1.9 GPU days to search cells with randomly selected channels, faster than computing all channels (3.3 GPU days). The method of randomly selecting channels performs a more efficient search without compromising the performance. In this way, we can also search architecture with larger batch sizes.

\subsubsection{Effects of joint supervision}

The batch triplet loss and center loss make use of feature maps before model prediction. The model under the joint supervision of classification, batch hard triplet, and center loss outperforms the model only with classificant loss by a significant margin, confirming the advantage of the joint loss function. As for the batch triplet loss, it obtains +0.7\%/+0.9\%/+0.2\% improvements on the precision/success/normalization precision scores, respectively. It is obvious that the center loss plays a vital role due to considerable intra-class variations, promoting performance with 0.4\% on the success score and 1.3\% on the precision score.

\subsection{Comparisons with state-of-the-arts}

We construct a TIR-PT model that utilizes the final searched and retrained network architecture and compare it with the state-of-the-art methods, including DBF~\cite{dbf}, DiMP~\cite{dimp}, ECO-stir~\cite{ecostir}, ECO-deep~\cite{eco}, HSSNet~\cite{hssnet}, MCFTS~\cite{mcfts}, MDNet~\cite{mdnet}, MLSSNet~\cite{mlssnet}, MMNet~\cite{mmnet}, SiamFC~\cite{siamfc}, SiamTri~\cite{siamtri}, and TADT~\cite{tadt}, on the LSOTB-TIR~\cite{lsotb} and PTB-TIR~\cite{ptb} benchmark datasets. Comparison results are shown in Table \ref{tab:comp}, our method achieves competitive performance.

LSOTB-TIR~\cite{lsotb} is a more comprehensive benchmark dataset due to the more intra-class variations of pedestrians in more challenging scenarios, such as background clutter, deformation, motion blur, and occlusion. It is evident that our method achieves the best performance compared with other state-of-the-art methods, the precision/success/normalized precision scores reach 0.805/0.669/0.720, which improves the baseline method, DiMP~\cite{dimp}, by 2.5\%/5.3\%/1.7\%, respectively. Compared with MMNet~\cite{mmnet}, which learns dual-level deep representation for TIR tracking, our method has more than 18\% improvement of all metrics on LSOTB-TIR.

The TIR-PT model searched and retrained on the LSOTB-TIR dataset can be indeed transferable to the PTB-TIR~\cite{ptb} dataset, achieves the best success score of 0.641 and a comparable precision score of 0.776, outperforming DiMP with relative gains of 2.7\% and 2.3\%, respectively. ECO-STIR~\cite{ecostir} uses ResNet-50 as its base network and utilizes synthetic TIR data generated from RGB data to train the model, which obtains the best precision score of 0.830, with a relative gain of 5.4\% compared to our method. Still, its success score is inferior to ours, with a degradation of 2.4\%. We also conduct evaluations on challenge scenarios often occurring in pedestrian tracking, as shown in Fig.\ref{fig:lsotb}. Our method achieves the competitive performance on all these challenges.

The above experimental result proves that our method can search a more effective and efficient network architecture for the TIR-PT task.

\begin{figure}[t!]
    \begin{center}
        \includegraphics[width=\linewidth]{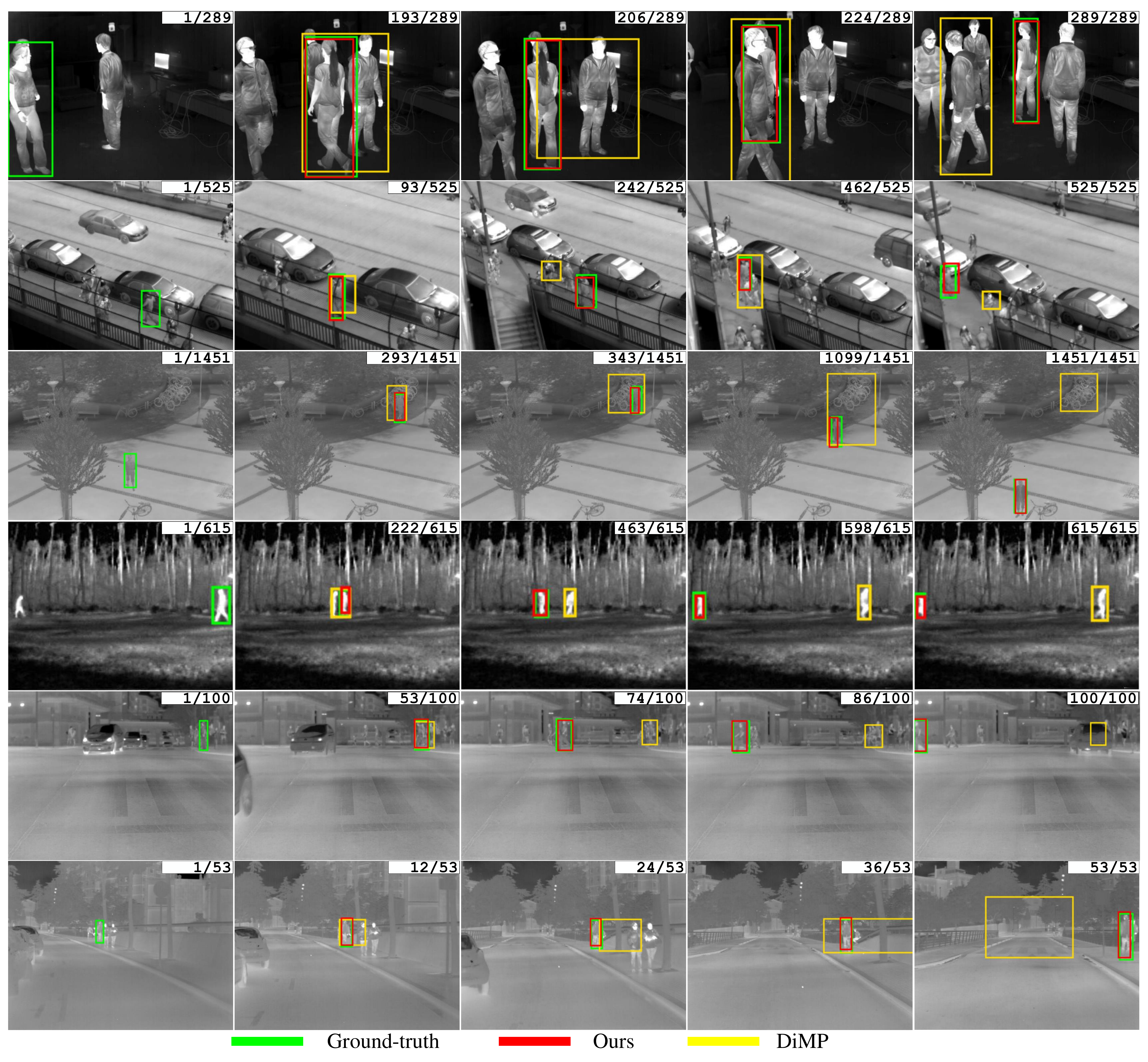}
    \end{center}
    \caption{Visualized comparison results of our method with the baseline DiMP tracker on six challenging video sequences from the PTB-TIR benchmark datasets. From top to bottom are \emph{classroom2}, \emph{crowd3}, \emph{jacket}, \emph{meetion3}, \emph{sidewalk3}, and \emph{street3}, respectively.}
    \label{fig:visual}
\end{figure}

\begin{figure}[t!]
    \begin{center}
        \includegraphics[width=0.5\linewidth]{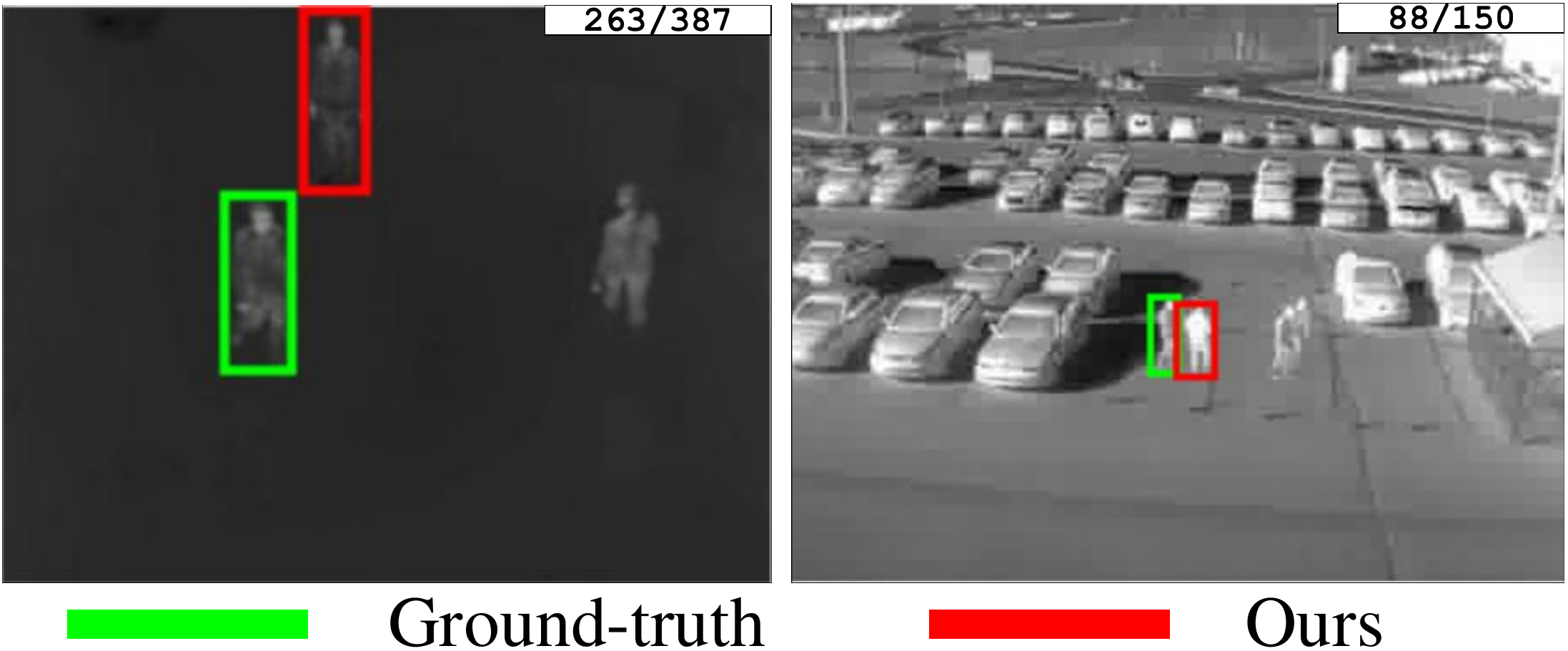}
    \end{center}
    \caption{Examples of the failed cases. From left to right are \emph{circle1} and \emph{distractor1}, respectively.}
    \label{fig:failure}
\end{figure}

\subsection{Visualization results}

In Fig.\ref{fig:visual}, we show the visualized comparison results of our method with the baseline DiMP tracker on six challenging video sequences, including \emph{classroom2}, \emph{crowd3}, \emph{jacket}, \emph{meetion3}, \emph{sidewalk3}, and \emph{street3}, from the PTB-TIR benchmark datasets, which utilizes ResNet-50 as its backbone network. Our method is more robust in scenes with deformation and fast motion. In comparison, DiMP is prone to be confused by background clutter and motion blur in these videos. These results indicate that the searched network architecture can learn more discriminative and robust representation. Figure \ref{fig:failure} illustrates two instances of tracking failures. One failure results from severe occlusion by the surrounding environment, causing the backbone network to lose track of the target object and fail to recover once the occlusion ends. This suggests that our current occlusion handling mechanism is insufficient when the occlusion entirely covers the object. Severe occlusions obscure the target object's distinctive features, complicating the tracker's ability to maintain its trajectory. Additionally, our method struggles with objects that have similar appearances and shapes. This problem occurs because the DiMP model predictor has difficulty differentiating the target object from visually similar distractors, leading to erroneous tracking. Future work will aim to enhance occlusion and distractor handling capabilities by incorporating more advanced operational candidates.

\section{Conclusion}\label{sec:5}

In this paper, we employ a NAS approach to find an effective network architecture for the TIR-PT task. Our design includes single-bottom and dual-bottom cells as fundamental search units, with a strategy of random channel selection to reduce computational costs and memory usage before feeding them into operation candidates. To enhance feature discrimination, we incorporate joint supervision using classification, batch hard triplet, and center loss. Despite requiring fewer computations, our method surpasses state-of-the-art approaches in overall performance. Extensive experiments validate the effectiveness and efficiency of our proposed method. However, as our method represents an initial effort in searching for efficient network architectures for TIR-PT, the simplicity of the basic search units limits the ability to capture target-specific spatiotemporal information. Future work will focus on integrating advanced attention mechanisms from other computer vision tasks into the basic search units, aiming to learn more discriminative and informative representations for improved TIR-PT performance.

\bibliographystyle{num}
\bibliography{bibliography}

\end{document}